\pdfoutput=1
\documentclass[11pt]{article}
\usepackage{ACL2023}
\usepackage{times}
\usepackage{latexsym}
\usepackage[T1]{fontenc}
\usepackage[utf8]{inputenc}
\usepackage{microtype}
\usepackage{graphicx}
\usepackage{censor}
\usepackage{inconsolata}
\usepackage{tabularx}
\usepackage{booktabs}
\usepackage{soul}
\usepackage{scalefnt}
\usepackage{etoolbox}
\patchcmd{\quote}{\rightmargin}{\leftmargin 1.8em \rightmargin}{}{}
\usepackage[inline]{enumitem}
\newlist{compactenum}{enumerate}{4}
\setlist[compactenum,1]{nolistsep,label=(\arabic*),leftmargin=6mm}
\newlist{compactitem}{itemize}{4}
\setlist[compactitem,1]{nolistsep,label=$\bullet$}
\makeatletter
\ifacl@finalcopy

\fi
\makeatother

\title{
  Where are We in Event-centric Emotion Analysis?\\
  Bridging Emotion Role Labeling and Appraisal-based Approaches
}

\author{Roman Klinger \\
  Institut f\"ur Maschinelle Sprachverarbeitung\\
  University of Stuttgart, Germany \\
  \texttt{roman.klinger@ims.uni-stuttgart.de}
}

\begin{document}
\maketitle
\begin{abstract}
  The term emotion analysis in text subsumes various natural language
  processing tasks which have in common the goal to enable computers
  to understand emotions. Most popular is emotion classification in
  which one or multiple emotions are assigned to a predefined textual
  unit. While such setting is appropriate for identifying the reader's
  or author's emotion, emotion role labeling adds the perspective of
  mentioned entities and extracts text spans that correspond to the
  emotion cause. The underlying emotion theories agree on one
  important point; that an emotion is caused by some internal or
  external event and comprises several subcomponents, including the
  subjective feeling and a cognitive evaluation. We therefore argue
  that emotions and events are related in two ways. (1)~Emotions are
  events; and this perspective is the fundament in natural language
  processing for emotion role labeling. (2)~Emotions are caused by
  events; a perspective that is made explicit with research how to
  incorporate psychological appraisal theories in NLP models to
  interpret events. These two research directions, role labeling and
  (event-focused) emotion classification, have by and large been
  tackled separately. In this paper, we contextualize both
  perspectives and discuss open research questions.
\end{abstract}

\begin{figure*}[t]
  \centering
  \includegraphics[width=0.999\linewidth]{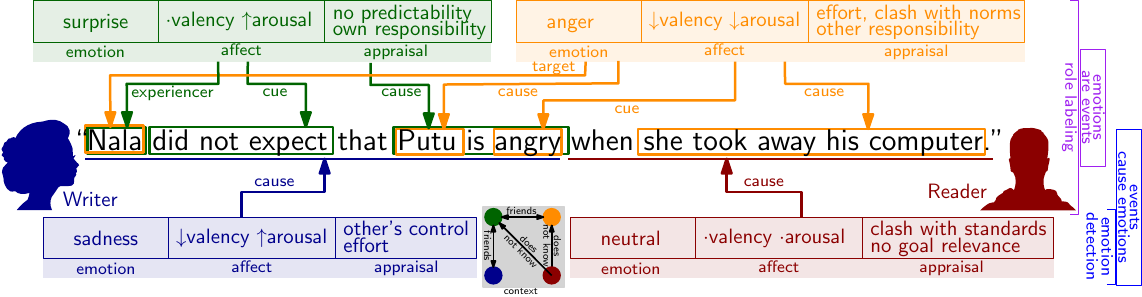}
  \caption{Integrated Visualization of Research Tasks in Emotion
    Analysis}
  \label{fig:setup}
\end{figure*}

\section{Introduction}
\begin{quote}\small
  ``Communication is an exchange of facts, ideas, opinions, or
  emotions by two or more persons. The exchange is successful only
  when mutual understanding results.''
  \citep[p. 219]{NewmanSummer1967}
\end{quote}
The development of computational models in natural language processing
aims at supporting communication between computers and humans; with
language understanding research focusing on enabling the computer to
comprehend the meaning of text. Sometimes, understanding facts is
sufficient, for instance when scientific text is analyzed to
automatically augment a database
\citep{li-etal-2016-commonsense,Trouillon2017}. Factual statements can
also comprise explicit reports of emotions or sentiments, such as
``They were sad.'', and in such cases, the analysis of subjective
language blends with information extraction
\citep{wiebe-etal-2004-learning}.

Emotion analysis, however, goes beyond such analysis of propositional
statements.  To better understand what emotion analysis models are
expected to do, it is worth reviewing emotion theories in
psychology. There are many of them, with varying purposes and
approaches, but most of them, if not all, agree on the aspect that
\textit{emotions are caused by some event} and come with a change of
various subsystems, such as a change in motivation, a subjective
perception, an expression, and bodily symptoms. Another component is
the evaluation of the causing event, sometimes even considered to
constitute the emotion \citep{Scarantino2016}.

The \textit{emotion also corresponds to an event itself}, embedded in
a context of other events, people, and objects. All components of such
emotion events (cause, stances towards other involved people, opinions
about objects) may be described along an explicit mention of an
emotion name. Any subset of them may appear in text, and may or may
not be sufficient to reliably assign an emotion representation to the
text author, a mentioned entity, or to a reader
\citep{Casel2021,cortal-etal-2023-emotion}.

This complexity has led to a set of various emotion analysis tasks in
NLP, which we exemplify in an integrated manner in
Figure~\ref{fig:setup}. The most popular task is emotion
prediction, either representing the writer's or the reader's
emotion as a category, as valence/arousal values, or as appraisal
vector (at the bottom of Figure~\ref{fig:setup}, we will describe the
underlying psychological theories in
\S\ref{sec:emotiontheories}).  Adding the task of cause
detection bridges to the role labeling setup (visualized in more
completeness at the top). Here, the emotion event is represented by
the token span that represents the emotion experiencer, the cue, and
the cause. \textit{\textcolor{darkblue}{Emotion prediction focuses on understanding
  from text how events cause emotions}, \textcolor{violet}{while role labeling focuses on
  understanding how emotions are represented as events themselves}.}

sWe now introduce the background to emotion analysis, including
psychological theories, related tasks, and use cases
(\S\ref{sec:relatedwork}). Based on that, we consolidate recent
research on the interpretation of events to infer an emotion and on
emotion role labeling (\S\ref{sec:appraisal}--\ref{sec:roles}). We
then point out existing efforts on bridging both fields
(\S\ref{sec:both}) and, based on this, develop a list of open research
questions (\S\ref{sec:futurework}). We show a visualization how
various NLP tasks and research areas are connected to emotion analysis
in Figure~\ref{fig:sketch} in the Appendix.

\section{Related Work}
\label{sec:relatedwork}
\subsection{Emotion Theories in Psychology}
\label{sec:emotiontheories}
Before we can discuss emotion analysis, we need to introduce what an
emotion is. The term
typically
refers to some feeling, some sensation,
that is defined following various perspectives. \citet{Scarantino2016}
provides an overview of various emotion theories and differentiates
between a \textit{motivation tradition}, a \textit{feeling tradition},
and an \textit{evaluative tradition}.

\subsubsection{Categorical Models of Basic Emotions} The motivation
tradition includes theories that are popular in NLP such as the basic
emotions proposed by \citet{Ekman1992} and \citet{Plutchik2001}. They
differ in how they define what makes an emotion basic: Ekman proposes
a list of properties, including an automatic appraisal, quick onset,
brief duration, and distinctive universal signals. According to him,
non-basic emotions do not exist but are rather emotional plots, moods,
or personality traits. Plutchik defines basic emotions based on their
function, and non basic-emotions are gradations and mixtures. The set
of basic emotions according to Ekman is commonly understood to
correspond to joy, anger, disgust, fear, sadness, and
surprise. However, in fact, the set is larger and there are even
emotions for which it is not yet known if they could be considered
basic \citep[e.g., relief, guilt, or love,][]{Ekman2011}. The basic
emotions according to Plutchik include anticipation and trust in
addition. In NLP, such theories mostly serve as a source for label
sets for which some evidence exists that they should be
distinguishable, also in textual analysis. A study that uses a
comparably large set of emotions is \citet{Demszky2020}, while many
other resource creation and modeling attempts focus on subsets
\citep[i.a.]{alm-etal-2005-emotions,strapparava-mihalcea-2007-semeval,schuff-etal-2017-annotation,li-etal-2017-dailydialog,mohammad-2012-emotional}.

\subsubsection{Dimensional Models of Affect} An alternative to
representing emotions as categorical labels is to place them in a
(continuous) vector space, in which the dimensions correspond to some
other meaning. The most popular one is the valence/arousal space, in
which emotions are situated according to their subjective perception
of a level of activation (arousal) and how positive the experience is
(valence). This concept stems from the feeling tradition mentioned
above and corresponds to affect \citep{Posner2005}. It also plays an
important role in constructionist theories, which aim at explaining
how the objectively measurable variables of valence and arousal may be
linked by cognitive processes to emotion categorizations
\citep{Barrett2017a}. While we are not aware of any applications of
the constructionist theories in NLP, emotion analysis has been
formulated as valence/arousal regression
\citep[i.a.]{Buechel2017,preotiuc-pietro-etal-2016-modelling}. Valence
and arousal predictions are related to, but not the same as, emotion
intensity regression \citep{Mohammad2017}.

\subsubsection{Appraisals} Affect is not the only so-called
dimensional model to represent emotions. More recently, the concept of
appraisals that represents the cognitive dimension of emotions, i.e.,
the cognitive evaluation of the event regarding the impact on the
self, found attention in NLP. The set of appraisals that can explain
emotions is not fixed and depends on the theory and the domain. It
often includes variables that describe if an event can be expected to
increase a required effort (likely to be high for anger or fear) or
how much responsibility the experiencer of the emotion holds (high for
feeling pride or guilt). \citet{Smith1985} showed that a comparably
small set of 6 appraisal variables can characterize differences
between 15 emotion categories. \citet{Scherer2001a} describes a
multi-step process of appraisal evaluations as one part of the emotion
-- their emotion component process model also reflects on additional
emotion components, namely the bodily reaction, the expression, the
motivational aspect, and the subjective feeling. Appraisal theories
led to a set of knowledge bases and models that link events to
emotions
\citep{Balahur2012,cambria-etal-2022-senticnet,Shaikh2009,Udochukwu2015},
but only recently, resources and models have been proposed which make
appraisal variables explicit
\citep{Stranisci2022,hofmann-etal-2020-appraisal,hofmann-etal-2021-emotion,troiano-etal-2022-x,Troiano2023,Wegge2022}.
This paper discusses work on appraisal theories to interpret events
regarding the potentially resulting emotion in
\S\ref{sec:appraisal}.

\subsection{Tasks Related to Emotion Analysis}
Emotion analysis is a task grounded in
various previous research fields, from which we discuss sentiment
analysis and personality profiling.

\begin{figure}
  \centering
  \includegraphics[width=0.85\linewidth]{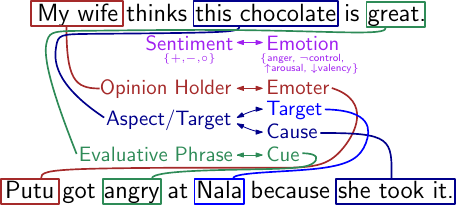}
  \caption{Comparison of structured sentiment analysis and emotion
    role labeling.}
  \label{fig:sentiment}
\end{figure}

\subsubsection{Sentiment Analysis} 
Sometimes, sentiment analysis is considered a simplified version of
emotion analysis in which multiple emotion categories are conflated
into two (positive or negative, sometimes distinguishing multiple
levels of intensity, \citet{Kiritchenko2016}). We would like to argue
that the tasks differ in more than the number of labels. Sentiment
analysis is often equated to classifying the text into a more
unspecific connotation of being positive or negative
\citep{liu2012sentiment}. Commonly, the sentiment of the text author
is analyzed, which renders the task to be overlapping with opinion
mining \citep{pang_opinion_2008,Barnes2017}. Emotion analysis is
hardly ever about detecting the opinion regarding a product; while
that is a common focus in sentiment analysis
\citep{pontiki-etal-2014-semeval}.

A more powerful approach to sentiment analysis is to not only detect
if the author expresses something positive, but also to detect opinion
holders, evaluated targets/aspects, and the phrase that describes the
evaluation
\citep{Barnes2022,pontiki-etal-2015-semeval,pontiki-etal-2016-semeval,klinger-cimiano:2013:Short}. The
tasks of such ``sentiment role labeling'' and ``emotion role
labeling'' do, however, barely match (see Figure~\ref{fig:sentiment}):
\begin{compactenum}
\item The \textit{opinion holder} in sentiment analysis is a person
  that expresses an opinion, regarding some object, service, or
  person. This commonly follows a cognitive evaluation, likely to be a
  conscious process rather than an unbidden reaction. We would therefore
  not call the person experiencing an emotion a ``holder'' but rather
  an \textit{emotion experiencer}, or \textit{feeler}, or an \textit{emoter} (to make the difference
  between an emotion and a feeling explicit).
\item The \textit{aspect/target} in sentiment analysis might
  correspond to two things in emotion analysis. It can be a
  \textit{target}, I can be angry \textit{at} someone, who is not
  solely the \textit{cause} of that emotion. I can be angry at a
  friend, because she did eat my emergency supply of chocolate. But I
  cannot be \textit{sad at} somebody. In emotion analysis, we care
  more about the \textit{stimulus} or \textit{cause} of an
  emotion. Sometimes, targets and causes are conflated.
\item The \textit{evaluative, subjective phrase} in sentiment analysis
  corresponds to emotion words (\textit{cue} in
  Figure~\ref{fig:setup}).
\end{compactenum}

It is noteworthy that evaluative statements in sentiment also express
an appraisal of something but the overlap with appraisal theories in
emotion analysis is minimal -- the evaluation of a product in
sentiment analysis is often expressed explicitly. On the contrary,
appraisal-based emotion analysis focuses on inferring the internal
appraisal processes of a person purely from an event description. We
refer the interested reader to
\newcite{martin_language_2005} for a comprehensive analysis of the
language used to describe evaluations.

\begin{figure}
  \centering
  \includegraphics[width=0.8\linewidth]{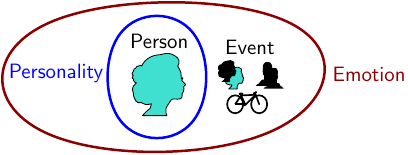}
  \caption{Comparison of personality detection and emotion analysis.}
  \label{fig:personality}
\end{figure}

\subsubsection{Personality Profiling} Sometimes the task of
personality analysis is seen to be similar to emotion analysis,
because both an emotion and the personality are based on a
person. Personality is, however, a function that depends only on the
person, while an emotion depends on the person in interaction with a
situation (see Figure~\ref{fig:personality}). Therefore, personality
is a stable trait, while emotions are states that change more flexibly
\citep{geiser_states_2017}. The most prominent model that found
application in NLP is the \textsc{Ocean}/Big-Five model
\citep{goldberg1999broad,roccas2002big}, comprising openness to
experience, conscientiousness, extraversion, agreeableness, and
neuroticism
\citep{pizzolli-strapparava-2019-personality,lynn-etal-2020-hierarchical,kreuter-etal-2022-items,golbeck2011predicting}. An
alternative is \textsc{Hexaco}, adding the dimension of honesty
\citep{lee2018psychometric}, which did, however, lead to less
attention in NLP \citep{sinha-etal-2015-mining}. Early work in
personality analysis based on linguistic features was based, similar
to sentiment or emotion analysis, on word-counting approaches
\citep{Pennebaker1999}.
The Myers–Briggs Type Indicator \citep[MBTI,][]{Myers1998} received
attention in NLP, partially because of a straight-forward way to
collect data with hash-tag-based self-supervision
\citep{plank-hovy-2015-personality,verhoeven-etal-2016-twisty}. This
model has weaknesses regarding reliability and validity
\citep{boyle_myers-briggs_1995,Randall2017} which affect the
robustness of NLP models \citep{stajner-yenikent-2021-mbti}.

\subsection{Use-Cases of Emotion Analysis}
Every kind of text in which an interpretation of the emotional
connotation is of value constitutes a potential use case for emotion
modeling. This includes the analysis of social media
\citep[i.a.]{Mohammad2018,Klinger2018,wang_harnessing_2012}, of news
articles \citep[i.a.]{Bostan2020}, of figurative language
\citep[i.a.]{chauhan-etal-2020-sentiment,dankers-etal-2019-modelling},
of abusive language
\citep[i.a.]{rajamanickam-etal-2020-joint,Plazadelarco2021} of
literature
\citep[i.a.]{Kim2018,Alm2005,Dodds-2011-Hedonometrics,Kim2017a}, of
clinically relevant disorders
\citep[i.a.]{Islam2018-hc,Pestian2012-su}, or the support of customer
agents \citep{labat-etal-2022-emotional}.

Each domain implicitly defines which subtasks are relevant. For news
headlines, the author's emotion is least interesting while estimating
the (intended) impact on the reader is important, for instance to
understand reactions in the society and intentional use to manipulate
readers \citep{caiani_dicocco_2023}. For hate speech detection or
other social media analysis tasks, the author's emotion is central. In
literature, an interesting aspect is to understand which emotion is
attributed to fictional characters
\citep{Kim2019,hoorn_perceiving_2003}.

Each domain also comes with particular challenges, stemming from
varying task formulations: News headlines are short and highly
contextualized in the outlet, the time of publication, and the reader's
stance towards topics \citep{schaffer_shocking_1995}. Social media
comes in informal language \citep{kern_gaining_2016}. Literature often
requires interpretations of longer text spans
\citep{kuhn_computational_2019}.
Each of these applications therefore comes with design choices:
\begin{compactitem}
\item What is the emotion perspective?\\ (reader, writer, entities)
\item What is the unit of analysis?\\ (headline, tweet, paragraph, $n$
  sentences)
\item Is text classification of predefined units sufficient or does a
  model need to assign emotions to automatically detected segments in
  the text?
\item What are the variables to be predicted and the possible value
  domain?\\ (emotion categories, appraisals, affect, spans of different
  kind)
\end{compactitem}

So far, models have mostly been developed for specific use-cases,
where such constraints can be clearly identified. This has, however,
an impact on the generalizability of models. We will now discuss the
two perspectives of \textit{events that cause emotions} as an
interpretation of emotion analysis as text classification of
predefined textual units (\S\ref{sec:appraisal}) and of \textit{events
  as emotions}, the case of emotion role labeling
(\S\ref{sec:roles}). After that, we explain the efforts to bring these
two directions together (\S\ref{sec:both}) and we build on top of this
consolidation to point out important future research directions
(\S\ref{sec:futurework}).

\section{The Link between Emotions and Events}
\subsection{Events cause Emotions: Appraisals}
\label{sec:appraisal}
\subsubsection{Traditional Emotion Analysis Systems}
Most emotion analysis systems were, before the deep learning
revolution in NLP, feature-based, and features often stemmed from
manually created lexicons \citep{Mohammad2013} and included manually
designed features for the task
\citep{StajnerKlinger2023,aman-szpakowicz-2007-identifying}. Since the
state of the art for the development of text analysis systems is
transfer learning by fine-tuning pretrained large language models
\citep[such as \textsc{Bert},][]{devlin-etal-2019-bert}, the
phenomenon-specific model development focuses on exploiting properties
of the concept. One example is DeepMoji, which adapts transfer
learning to the analysis of subjective language and identifies a
particularly useful pretraining task, namely the prediction of emojis
\citep{felbo-etal-2017-using}. Another strain of research aims at
developing models that aggregate multiple emotion theories
\citep{buechel-etal-2021-towards}.

\begin{figure}
  \centering\tiny\sffamily\scalefont{0.99}
  \newcommand{\ittrk}{\raisebox{.5pt}{\scalebox{0.6}{$\bullet$}}\hspace{-4pt}}
    \begin{tabular}{|p{.19\linewidth}|p{0.21\linewidth}|p{.17\linewidth}|p{.22\linewidth}|}\hline
              &             &        & Normative \\[-.5mm]
    Relevance & Implication & Coping & Significance\\
    \hline
    \parbox[t][3.6cm]{\linewidth}{%
    \ul{Novelty}
    \begin{compactitem}[nolistsep,label={\ittrk},leftmargin=0mm]
    \item suddenness
    \item familiarity
    \item predictability
    \end{compactitem}
    \begin{compactitem}[nolistsep,label={\ittrk},leftmargin=0mm]
    \item attention
    \item att. removal
    \end{compactitem}
    \vspace{0.5\baselineskip}
    \ul{Intrinsic\\ Pleasantness}
    \begin{compactitem}[nolistsep,label={\ittrk},leftmargin=0mm]
    \item pleasant
    \item unpleasant
    \end{compactitem}
    \vspace{0.5\baselineskip}
    \ul{Goal Relevance}
    \begin{compactitem}[nolistsep,label={\ittrk},leftmargin=0mm]
    \item goal-related
    \end{compactitem}
    }
              &
    \parbox[t][2cm]{\linewidth}{%
    \ul{Causality: agent}
    \begin{compactitem}[nolistsep,label={\ittrk},leftmargin=0mm]
    \item own responsib.
    \item other's respons.
    \item situational resp.
    \end{compactitem}
   \vspace{0.5\baselineskip}
   \ul{Goal\\ conduciveness}
   \begin{compactitem}[nolistsep,label={\ittrk},leftmargin=0mm]
   \item goal support
   \end{compactitem}
   \vspace{0.5\baselineskip}
   \ul{Outcome\\ probability}
   \begin{compactitem}[nolistsep,label={\ittrk},leftmargin=0mm]
   \item consequence\\ anticipation
   \end{compactitem}
   \vspace{0.5\baselineskip}
   \ul{Urgency}
   \begin{compactitem}[nolistsep,label={\ittrk},leftmargin=0mm]
   \item response urgency
   \end{compactitem}
   \vspace{0.5\baselineskip}
   }

&
   \parbox[t][2cm]{\linewidth}{%
   \ul{Control}
   \begin{compactitem}[nolistsep,label={\ittrk},leftmargin=0mm]
   \item own control
   \item others' control
   \item chance\\ control
   \end{compactitem}
   \vspace{0.5\baselineskip}
   \ul{Adjustment}
   \begin{compactitem}[nolistsep,label={\ittrk},leftmargin=0mm]
   \item anticipated\\ acceptance
   \end{compactitem}
   \begin{compactitem}[nolistsep,label={\ittrk},leftmargin=0mm]
   \item effort
   \end{compactitem}
  \vspace{0.5\baselineskip}
  }
         &
   \parbox[t][2cm]{\linewidth}{%
   \ul{Internal standards compatibility}
   \begin{compactitem}[nolistsep,label={\ittrk},leftmargin=0mm]
   \item clash with own standards
   \end{compactitem}
    \vspace{0.5\baselineskip}
    \ul{External standards compatibility}
    \begin{compactitem}[nolistsep,label={\ittrk},leftmargin=0mm]
    \item clash with norms
    \end{compactitem}
    \vspace{0.5\baselineskip}
   }
    \\
    \hline
  \end{tabular}

  \caption{Variables used by \citet{Troiano2023} to analyze text
    according to combined dimensions proposed by 
    \citet{Scherer2001a} and \citet{Smith1985}.\\[-20pt]}
  \label{fig:sequential}
\end{figure}

\subsubsection{Event Interpretation} We focus on the aspect of
emotions that they are caused by events. Interpreting events is
challenging, because event descriptions often lack an explicit emotion
mention \citep{Troiano2023a}. Such textual instances are considered
``implicit'' regarding their emotion
\citep{Udochukwu2015,Klinger2018}: The challenge to be solved is to
link ``non-emotional'' events to the emotion that they might
cause. \citet{Balahur2012} tackled this by listing action units in an
ontology, based on semantic parsing of large amounts of text.
\citet{cambria-etal-2022-senticnet} developed a logics-based resource
to associate events with their emotion interpretation.

\subsubsection{Incorporating Appraisal Variables in Text Analysis Models} These
attempts, however, do not model appraisal variables explicitly as a
link between cognitive evaluations of events and emotions. There is
also not only one appraisal theory, and depending on the theory, the
computational modeling is realized in differing ways. Based on the OCC
model \citep[an appraisal theory that provides a decision tree of
appraisal variables to characterize emotions,][]{Steunebrink2009},
both \citet{Shaikh2009} and \citet{Udochukwu2015} develop methods to
extract atomic variable values from text that are the building blocks
for appraisal-based interpretations. An example appraisal variable is
if an event is directed towards the self, for which they use semantic
and syntactic parsers. Other such variables include the valence of
events, the attitude towards objects, or the moral evaluation of
people's behaviours -- all detected with polarity lexicons. These
variables are then put together with logical rules, such as \texttt{If
  Direction = `Self' and Tense = `Future' and Overall Polarity =
  `Positive' and Event Polarity = `Positive', then Emotion = `Hope'}
\citep{Udochukwu2015}. The advantage of this approach is that it makes
the appraisal-based interpretation explicit; however, it does not
allow for reasoning under uncertainty, partially because these studies
do not build on top of manually assigned appraisal variables to text.

\subsubsection{Appraisal-Annotated Corpora} To understand the link better
between appraisals in text and emotions,
\citet{hofmann-etal-2020-appraisal} manually annotated
autobiographical event reports \citep{troiano-etal-2019-crowdsourcing}
for the appraisal dimensions identified by \citet{Smith1985}: does the
writer want to devote attention, were they certain about what was
happening, did they have to expend mental or physical effort to deal
with the situation, did they find the event pleasant, were they
responsible for the situation, could they control the situation, and
did they find that the situation could not be changed by anyone? They
found that the annotation replicates the links to emotions as found in
original studies \citep[Fig.~1]{hofmann-etal-2021-emotion}. Further,
they showed that appraisals can reliably be detected, but they did not
manage to develop a model that predicts emotions better with the help
of appraisals than without. Hence, they proposed a new way of modeling
emotions in text, but did not succeed to develop a multi-emotion model.

\begin{figure}
  \centering
\includegraphics[width=1.01\linewidth]{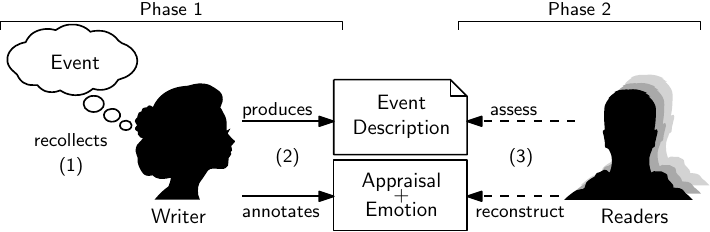}
  \caption{The study design that lead to the 
    crowd-enVENT data set \citep{Troiano2023}.\\[-26pt]}
  \label{fig:crowdenVent}
\end{figure}

\subsubsection{Appraisal Annotations by Event Experiencers} To
understand better if this inferiority of a joint model might be a
result of an imperfect noisy appraisal annotation, and to create a
larger corpus, \citet{Troiano2023} setup the experiment depicted in
Figure~\ref{fig:crowdenVent} (replicating
\citet{troiano-etal-2019-crowdsourcing}, but with appraisal
variables). They asked crowdworkers to describe an event that caused a
specific emotion and to then assign appraisal values \citep[this time
following the sequential approach by][with 21 variables,
Figure~\ref{fig:sequential}]{Scherer2001a} how they perceived the
respective situation (Phase 1). They then asked other people to read
the texts and reconstruct the emotion and appraisal (Phase
2). Unsurprisingly, the readers sometimes misinterpreted an event. For
instance ``I put together a funeral service for my Aunt'' is mostly
interpreted as something sad, while the original author was actually
proud about it. These differences in interpretation can also
 be seen in the appraisal variables --
Appraisals explain the differences in the event evaluation:
The interpretation as being sad comes with evaluations as not being in
control, while the interpretation to cause pride comes with being in
control.

\subsubsection{Emotion Modeling under Consideration of Appraisals}
The modeling experiments of \citet{Troiano2023} confirm that also a
larger set of variables can be reliably detected -- similarly well as
humans can reconstruct them. To further understand if such
self-assigned appraisal labels enable an improvement also in the
emotion categorization, they fine-tuned Ro\textsc{Bert}a
\citep{Liu2019Roberta} and tested if adding appraisal values improves
the result. They find that appraisals help the prediction of anger,
fear, joy, pride, guilt, sadness, and anger. They showcase the event
report ``His toenails were massive.'', where the baseline
%emotion
model relies on something massive being associated to pride. With the
appraisal information, it correctly assigns ``disgust''.

\subsubsection{Other Research Directions} More recently other research
has been published with a focused on specific
use-cases. \citet{Stranisci2022} who follow the appraisal model by
\citet{roseman_appraisal_2013} postannotate Reddit posts which deal
with situations that challenged the author to cope with an undesirable
situation. Their \textsc{App}Reddit corpus is the first resource of
appraisal-annotated texts from the
wild. \citet{cortal-etal-2023-emotion} follow a similar idea and
acquire texts that describe how people regulate their emotions in
specific situations. Next to their resource creation effort for
French, they analyze which descriptions of cognitive processes allow
to infer an emotion.

\medskip\noindent
We conclude that appraisal-based emotion analysis research has the
goal to better understand how emotions are implicitly communicated and
to develop better emotion analysis systems.

\subsection{Emotions are Events: Structured Analysis}
\label{sec:roles}
\begin{figure}
  \centering
  \includegraphics[width=\linewidth]{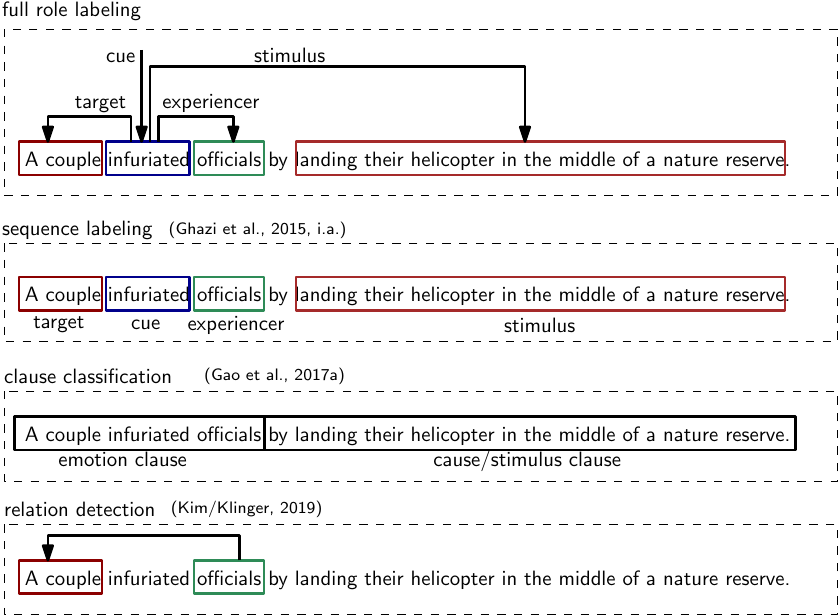}
  \caption{Emotion Role Modeling approaches (example from
    \citet{Bostan2020}). Full emotion role labeling has not been
    performed yet (top).}
  \label{fig:erlmodeling}
\end{figure}

The studies that we discussed so far put the aspect of emotion
analysis on the spot that emotions are caused by events. As we argued
before, emotions also constitute events. Similarly to the field of
semantic role labeling \citep{gildea-jurafsky-2000-automatic} which
models events in text following frame semantics, various efforts have
been made to extract emotion event representations from text. The
corpora that have been created come with differing modeling attempts,
summarized in Figure~\ref{fig:erlmodeling}.

\subsubsection{Cue Phrase Detection} The early work by
\citet{aman-szpakowicz-2007-identifying} focused on the emotion
\textit{cue} word, as an important part of role labeling.  They
annotated sentences from blogs, but did not propose an automatic cue
identification system. A structurally similar resource with cue word
annotations has been proposed by \citet{liew-etal-2016-emotweet}.

\subsubsection{Stimulus Detection} A few corpora have been developed
focussing on stimuli: \citet{ghazi2015} annotated sentences from
FrameNet that are known to be associated with emotions and model the
automatic prediction as sequence labeling.  For German,
\citet{DoanDang2021} created a similar corpus based on news
headlines. \citet{gao2017} formulated stimulus detection as clause
classification in Mandarin, which might, however, not be an
appropriate approach for English \citep{Oberlaender2020b}.

\subsubsection{Role Labeling as Classification} An interesting attempt
of emotion role labeling in texts from social media was the study on
Tweets associated to a US election by \citet{Mohammad2014}. The
decision to focus on a narrow domain allowed them to frame the role
identification task both in crowdsourced annotation and in modeling as
a classification task; namely to decide if the emoter, the stimulus or
the emotion target correspond to an entity from a predefined set (this
modeling formulation is not shown in Figure~\ref{fig:erlmodeling}).

\subsubsection{Full Emotion Role Labeling Resources} \citet{Kim2018}
and \citet{Bostan2020} aimed at creating corpora with full emotion
role labeling information. The \textsc{Reman} corpus \citep{Kim2019}
focused on literature from Project Gutenberg. Given the challenging
domain, the authors decided to carefully train annotators instead of
relying on crowdsourcing. Each instance corresponds to a sentence
triple, in which the middle sentence contains the cue to which the
roles of emoters, targets, and stimuli are to be associated. The
sequence-labeling-based modeling revealed that cause and target
detection are very challenging.  The paper does not contain an effort
to reconstruct the full emotion event graph structure.

\citet{Bostan2020} annotated news headlines, under the assumption that
less context is required for interpretation (which turned out to not
be true). To attribute for the subjective nature of emotion
interpretations, they setup the annotation as a multi-step
crowdsourcing task. The modeling experiments on their
\textsc{G}ood\textsc{N}ews\textsc{E}veryone corpus are limited to span
prediction.

\subsubsection{Role Labeling as Relation Detection} We are only aware of
one work in the context of semantic role labeling that attempts to
model the relational structure. \citet{Kim2019} simplified role
labeling to relation classification of emotional relations between
entities. This allowed them to build on top of established methods for
relation detection \citep{zhou-etal-2016-attention} but they
sacrificed explicit cue word detection and limited the analysis to
emotion stimuli that have a corresponding entity.

\subsubsection{Aggregated Corpora} There have been two efforts of data
aggregation, by \citet{Oberlaender2020b} and
\citet{Campagnano2022}. The latter compared various models for role
detection via span prediction. The prior we will discuss in the next
section. To sum up, there have been some efforts to perform emotion
role labeling, but in contrast to generic role labeling or to
structured sentiment analysis, no models have yet been developed for
full graph reconstruction. We visualize the differences in modeling
attempts in Figure~\ref{fig:erlmodeling}.

\begin{figure}
  \centering
  \includegraphics[width=\linewidth]{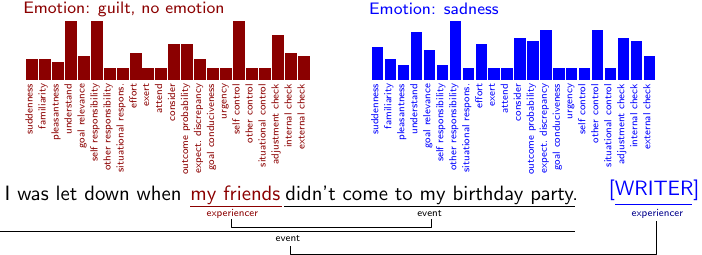}
  \caption{Example from the x-enVENT dataset}
  \label{fig:emorel}
\end{figure}
\subsection{Bridging the Two Perspectives}
\label{sec:both}
We now discussed the two perspectives of \textit{events causing
  emotions} (\S\ref{sec:appraisal}) and \textit{emotions being events}
(\S\ref{sec:roles}). The fact that these two analysis tasks have so
far mostly been tackled separately leaves a lot of space for future
research. However, some attempts to link the two areas already exist.

\subsubsection{Do the tasks of emotion classification and role labeling
  benefit from each other?} \citet{Oberlaender2020b} aimed at
understanding if knowledge of roles impacts the performance of emotion
categorization. It turns out it does, either because the relevant part
of the text is made more explicit (stimulus), or because of biases
(emoter).

Similarly, \citet{xia-ding-2019-emotion} setup the task of
stimulus-clause and emotion-clause pair classification. Their corpora
and a plethora of follow-up work show that stimulus and emotion
detection benefit from each other.

\subsubsection{Descriptions of which emotion components enable emotion
  recognition?} A similar strain of research aims at understanding
which components of emotions support emotion
predictions. \citet{Casel2021} performed multi-task learning
experiments with emotion categorization and emotion component
prediction. \citet{kim-klinger-2019-analysis} study how specific
emotions are communicated, similarly to
\citet{etienne-etal-2022-psycho}. \citet{cortal-etal-2023-emotion}
analyzed if particular ways of cognitively evaluating events support
the emotion prediction more than others.

\subsubsection{Linking Role Labeling and Appraisal-based Analysis} These
works do, however, not link emotion roles explicitly to their
cognitive evaluation dimensions. The only work that aimed at doing so
is the corpus by \citet{troiano-etal-2022-x}, who label emoters for
emotion categories and appraisals, the events that act as a stimulus
on the token level, and the relation between
them. Figure~\ref{fig:emorel} shows an example from their corpus. In
their modeling efforts, however, they limited themselves to
emoter-specific emotion/appraisal predictions and ignored, so far, the
span-based stimulus annotations \citep{Wegge2022,Wegge2023}.

\section{Open Research Tasks}
\label{sec:futurework}
We have now discussed previous work in emotion analysis,
appraisal-based approaches and role labeling. In the following, we
will make a set of aspects explicit that, from our perspective, need
future work.

\newcommand{\taskhead}[1]{\paragraph{#1.}}

\taskhead{Full emotion role labeling} Several corpora exist now that
have complex annotations of the emoter, their respective emotion
stimuli, targets, and cue words; partially with sentence level
annotations for the reader and writer in addition.  Modeling, however,
focused on sequence labeling for subsets of the roles or sentence
level classification. There are no attempts of full emotion graph
prediction, despite that role prediction subtasks might benefit from
being modeled jointly. There is also only little work on exploiting
role information for emotion categorization on the sentence level, a
potentially valuable approach for joint modeling of a structured
prediction task with text classification.

\taskhead{Role labeling/stimulus detection with appraisal information}
The work that has been performed to understand the interaction between
role prediction and emotion categorization focused on predicting
discrete emotion classes. However, stimuli often correspond to event
descriptions and therefore are a straight-forward choice for further
analysis with appraisal variables. Also, understanding which event
mentions in a text can function as an emotion stimulus could be
supported with the help of appraisals. The detection of clauses or
token sequences that correspond to emotion stimuli in context of
appraisal-based interpretations therefore has potential to improve
both subtasks.

\taskhead{Integration of other emotion models in role labeling}
Emotion categorization is typically one variable to be predicted in
stimulus detection and role labeling approaches, either for a writer
or for entities. An additionally interesting approach would be to
integrate other emotion representations with role labeling. An
interesting choice would be to create a corpus of valence/arousal
values, assigned to specific entities and linked to stimuli. Such
approach comes with the general advantage of dimensional models,
namely that emotion categories do not need to be predefined.

\taskhead{Robust cross-corpus modeling and zero-shot predictions} A
similar motivation lead to recent work on zero-shot emotion
prediction, in which emotion categories are to be predicted that are
not available in the training data. \citet{Plazadelarco2021} showed
that the performance loss of natural language inference-based
prompting in comparison to supervised learning leaves space for
improvements. Such attempts might also bridge the gap between
in-domain performance and cross-domain performance of emotion analysis
systems \citep{bostan-klinger-2018-analysis}. Zero-shot modeling or
other approaches to find representations that are agnostic to the
underlying emotion theory are essential for cross-corpus experiments,
because the domains that are represented by different corpora require
differing label sets.

\taskhead{Interpretation of event chains} Textual event descriptions
can be interpreted with appraisal theories, but we rely on end-to-end
learning to understand how sequences of events lead to specific
emotions (for instance being afraid of a specific unconfirmed
undesirable event $e$ $\rightarrow$ $e$ is disconfirmed $\rightarrow$
relief). Dissecting events with semantic parsing, and combining them
with emotion role labeling leads to sequences of general and emotion
events, which can be the input for a second-level emotion
analysis. Such methods would be required to fully understand how
emotions develop throughout longer sequences of stories, for instance
in literature.

\taskhead{Perspectivism} Appraisals do explain differences in the
emotion assessement, based on differing interpretations of events
\citep{Troiano2023}. We do, however, not know the role of underlying
factors. A perspectivistic approach with the goal to uncover variables
that lead to varying emotion constructions, e.g., based on demographic
data of event participants or other data, might provide additional
insight. This could also be applied to literature analysis, for
instance by including personality information on fictional characters
in the emotion prediction \citep{bamman-etal-2013-learning}. Such
approach is well-motivated in psychology; we know that personality
influences the interpretation of other's emotions
\citep{Doellinger2021}.

\taskhead{Integrate emotion models from psychology} Emotion analysis
work so far focused on a comparably small set of emotion theories. The
philosophical discussion by \citet{Scarantino2016} offers itself as a
guideing principle which other theories might be valueable to be
explored. This does not only include entirely so-far-ignored theories
\citep[e.g.,][]{Barrett2017a} but also knowledge from theories popular
in NLP. For instance, \citet{Ekman1992,Plutchik2001} offer more
information than lists of emotion categories. Integrating
psychological knowledge in NLP models can improve the performance
\citep{Troiano2023}. In a similar vein, there exist specific appraisal
theories for particular domains, including, e.g., argumentation theories
\cite{Dillard2012}.

\taskhead{Multimodal Modeling} We focused in our paper on analysis
tasks from text, but there has already been work on multimodal emotion
analysis \citep[i.a.]{Busso2008} and detecting emotion stimuli in
images \citep[i.a.]{Dellagiacoma2011,Fan2018}, also multimodally
\citep{khlyzova-etal-2022-complementarity,Cevher2019}. However, we are
not aware of any work in computer vision that interprets situations
and the interactions of events with the help of appraisal theories. To
fully grasp available information in everyday communication or
(social) media, the presented approaches from this paper need to be
extended multimodally.

\taskhead{Multilingual modeling} Most papers that we discuss in this
paper focus on English -- with very few exceptions, which we pointed
out explicitly. We are not aware of any emotion role labeling corpus
with full graph annotations in other languages, and there are only
very few attempts to integrate appraisal theories in emotion detection
on languages other than English. Such multilingual extension is not
only relevant to achieve models that work across use-cases -- the
concept of emotion names might also differ between languages, and
therefore comparing emotion concepts with the help of dimensional
appraisal models between languages and cultures can provide
interesting insights for both NLP and psychology.

\section{Conclusion}
With this paper, we discussed appraisal theory-based methods to
interpret events, and how emotions can be represented as events with
role labeling. We did that guided by our own two emotion analysis
projects SEAT (Structured Multi-Domain Emotion Analysis from Text) and
CEAT (Computational Event Evaluation based on Appraisal Theories for
Emotion Analysis) which corresponded each to one of the two
perspectives.

These two fields have been approached mostly separately
so far and the main goal of this paper is to make the research
narrative behind both transparent, and, based on this, point out open
research questions. Such open tasks emerge from missing connections
between the various goals in emotion analysis, but there are also
other promising directions that we pointed out.

We do not believe that this list is comprehensive, but hope that the
aggregation of previous work and pointing at missing research helps
interested researchers to identify the gaps they want to fill. Emotion
analysis is important to make computers aware of the concept, which is
essential for natural communication.

In addition, research in these fields helps to better understand how
humans communicate, beyond building impactful computational
systems. Therefore, research in affective computing brings together
psychology, linguistics, and NLP.

\section*{Limitations}
This paper focused on appraisal theories and emotion role labeling
mostly from a theoretical perspective. We aimed at pointing out open
research questions mostly based on conceptualizations of theories from
semantics and psychology. To identify open research questions, a
closer introspection of existing models need to be performed in
addition. In our theoretical discussion, we assume that the open
research questions have similar chances to succeed. In practical terms
this is likely not the case and we therefore propose to first perform
preliminary studies before definitely deciding to follow one of the
research plans that we sketched.

\section*{Ethics Statement}
The contributions in this paper do not directly pose any ethical
issues: we did not publish data, models, or did perform
experiments. However, the open topics that we identified might lead to
resources and models that can in principle do harm to
people. Following deontological ethics, we assume that no emotion
analysis systems should be applied to data created by a person without
their consent, if the results are used not only in aggregated form
which would allow to identify the person who is associated with the
analyzed data. We personally do not believe that a utilitaristic
approach may be acceptable in which reasons could exist that justify
to use emotion analysis technology to identify individuals from a
larger group. This is particularly important with methods discussed in
this paper in comparison to more general emotion categorization
methods, because we focus on implicit emotion expressions. The methods
we discussed and future work we sketched would be able to identify
emotions that are not explicitly expressed, and therefore humans that
generate data might not be aware that their private emotional state
could be reconstructed from the data they produce.

When creating data for emotion analysis, independent of its language,
domain, or the task formulation as role labeling, classification,
regression, using a dimensional model or a theory of basic emotions,
fairness or developed system and bias in data and systems is typically
an issue. While efforts exist to identify unwanted bias and
confounders in automatic analysis systems, the possible existance of
unidentified biases can never be excluded. Therefore, automatic
systems always need to be applied with care while critically
reflecting the automatically obtained results. This is particularly
the case with systems that focus on interpreting implicit emotion
communications that require reasoning under uncertainty. To enable
such critical reflection of a system's output, their decision must be
transparently communicated to the users.

In general, the ability of automatic systems to interpret and
aggregate emotions should not be used unaware of the people who
created data, and decisions and actions following recognized emotions
always need to remain in the responsibility of a human user.

We see our work mostly as a research contribution with the goal to
better understand how humans communicate, not as an automatic enabling
tool to provide insight in the private states of people.

\section*{Acknowledgements}
We would like to thank all coauthors who contributed to our work on
emotion analysis with the help of appraisal theories and in role
labeling. These are (in alphabetical order) Amelie Heindl, Antje
Schweitzer, Bao Minh Doan Dang, Enrica Troiano, Evgeny Kim, Felix
Casel, Flor Miriam Arco Del Plaza, Hendrik Schuff, Jan Hofmann, Jeremy
Barnes, Kai Sassenberg, Kevin Reich, Laura Oberl\"ander n\'ee Bostan,
Max Wegge, Sebastian Pad\'o, Tornike Tsereteli, and Valentino
Sabbatino. We further thank Alexandra Balahur, Orph\'ee De Clercq,
Saif Mohammad, Veronique Hoste, Valentin Barriere, and Sanja
\v{S}tajner for discussions on the general topics of emotion analysis
that helped us to develop this paper.

This work has been funded by two projects of the German Research
Council (Deutsche Forschungsgemeinschaft), namely the project
``Structured Multi-Domain Emotion Analysis from Text'' (SEAT, KL
2869/1-1) and ``Computational Event Evaluation based on Appraisal
Theories for Emotion Analysis'' (CEAT, KL
2869/1-2).\footnote{\url{https://www.ims.uni-stuttgart.de/en/research/projects/seat/},
  \url{https://www.ims.uni-stuttgart.de/en/research/projects/ceat/}}

\bibliography{lit}

\begin{thebibliography}{113}
\expandafter\ifx\csname natexlab\endcsname\relax\def\natexlab#1{#1}\fi

\bibitem[{Alm et~al.(2005)Alm, Roth, and Sproat}]{alm-etal-2005-emotions}
Cecilia~Ovesdotter Alm, Dan Roth, and Richard Sproat. 2005.
\newblock \href {https://aclanthology.org/H05-1073} {Emotions from text:
  Machine learning for text-based emotion prediction}.
\newblock In \emph{Proceedings of Human Language Technology Conference and
  Conference on Empirical Methods in Natural Language Processing}, pages
  579--586, Vancouver, British Columbia, Canada. Association for Computational
  Linguistics.

\bibitem[{Alm and Sproat(2005)}]{Alm2005}
Cecilia~Ovesdotter Alm and Richard Sproat. 2005.
\newblock \href {https://doi.org/10.1007/11573548_86} {Emotional sequencing and
  development in fairy tales}.
\newblock In \emph{Affective Computing and Intelligent Interaction}, pages
  668--674, Berlin, Heidelberg. Springer Berlin Heidelberg.

\bibitem[{Aman and Szpakowicz(2007)}]{aman-szpakowicz-2007-identifying}
Saima Aman and Stan Szpakowicz. 2007.
\newblock \href {https://doi.org/10.1007/978-3-540-74628-7_27} {Identifying
  expressions of emotion in text}.
\newblock In \emph{Text, Speech and Dialogue}, pages 196--205, Berlin,
  Heidelberg. Springer Berlin Heidelberg.

\bibitem[{Balahur et~al.(2012)Balahur, Hermida, and Montoyo}]{Balahur2012}
Alexandra Balahur, Jesus~M. Hermida, and Andrew Montoyo. 2012.
\newblock \href {https://doi.org/10.1109/T-AFFC.2011.33} {Building and
  exploiting emotinet, a knowledge base for emotion detection based on the
  appraisal theory model}.
\newblock \emph{IEEE Transactions on Affective Computing}, 3(1):88--101.

\bibitem[{Bamman et~al.(2013)Bamman, O{'}Connor, and
  Smith}]{bamman-etal-2013-learning}
David Bamman, Brendan O{'}Connor, and Noah~A. Smith. 2013.
\newblock \href {https://aclanthology.org/P13-1035} {Learning latent personas
  of film characters}.
\newblock In \emph{Proceedings of the 51st Annual Meeting of the Association
  for Computational Linguistics (Volume 1: Long Papers)}, pages 352--361,
  Sofia, Bulgaria. Association for Computational Linguistics.

\bibitem[{Barnes et~al.(2017)Barnes, Klinger, and Schulte~im
  Walde}]{Barnes2017}
Jeremy Barnes, Roman Klinger, and Sabine Schulte~im Walde. 2017.
\newblock \href {http://www.ims.uni-stuttgart.de/data/sota_sentiment}
  {Assessing state-of-the-art sentiment models on state-of-the-art sentiment
  datasets}.
\newblock In \emph{Proceedings of the 8th Workshop on Computational Approaches
  to Subjectivity, Sentiment and Social Media Analysis}, Copenhagen, Denmark.
  Workshop at Conference on Empirical Methods in Natural Language Processing,
  Association for Computational Linguistics.

\bibitem[{Barnes et~al.(2022)Barnes, Oberlaender, Troiano, Kutuzov, Buchmann,
  Agerri, {\O}vrelid, and Velldal}]{Barnes2022}
Jeremy Barnes, Laura Oberlaender, Enrica Troiano, Andrey Kutuzov, Jan Buchmann,
  Rodrigo Agerri, Lilja {\O}vrelid, and Erik Velldal. 2022.
\newblock \href {https://doi.org/10.18653/v1/2022.semeval-1.180} {{S}em{E}val
  2022 task 10: Structured sentiment analysis}.
\newblock In \emph{Proceedings of the 16th International Workshop on Semantic
  Evaluation (SemEval-2022)}, pages 1280--1295, Seattle, United States.
  Association for Computational Linguistics.

\bibitem[{Bostan et~al.(2020)Bostan, Kim, and Klinger}]{Bostan2020}
Laura Ana~Maria Bostan, Evgeny Kim, and Roman Klinger. 2020.
\newblock \href {https://www.aclanthology.org/2020.lrec-1.194}
  {{G}ood{N}ews{E}veryone: A corpus of news headlines annotated with emotions,
  semantic roles, and reader perception}.
\newblock In \emph{Proceedings of The 12th Language Resources and Evaluation
  Conference}, pages 1554--1566, Marseille, France. European Language Resources
  Association.

\bibitem[{Bostan and Klinger(2018)}]{bostan-klinger-2018-analysis}
Laura-Ana-Maria Bostan and Roman Klinger. 2018.
\newblock \href {https://aclanthology.org/C18-1179} {An analysis of annotated
  corpora for emotion classification in text}.
\newblock In \emph{Proceedings of the 27th International Conference on
  Computational Linguistics}, pages 2104--2119, Santa Fe, New Mexico, USA.
  Association for Computational Linguistics.

\bibitem[{Boyle(1995)}]{boyle_myers-briggs_1995}
Gregory~J. Boyle. 1995.
\newblock \href {https://doi.org/10.1111/j.1742-9544.1995.tb01750.x}
  {Myers-{Briggs} {Type} {Indicator} ({MBTI}): {Some} {Psychometric}
  {Limitations}}.
\newblock \emph{Australian Psychologist}, 30(1):71--74.

\bibitem[{Buechel and Hahn(2017)}]{Buechel2017}
Sven Buechel and Udo Hahn. 2017.
\newblock \href {https://aclanthology.org/E17-2092} {{E}mo{B}ank: Studying the
  impact of annotation perspective and representation format on dimensional
  emotion analysis}.
\newblock In \emph{Proceedings of the 15th Conference of the {E}uropean Chapter
  of the Association for Computational Linguistics: Volume 2, Short Papers},
  pages 578--585, Valencia, Spain. Association for Computational Linguistics.

\bibitem[{Buechel et~al.(2021)Buechel, Modersohn, and
  Hahn}]{buechel-etal-2021-towards}
Sven Buechel, Luise Modersohn, and Udo Hahn. 2021.
\newblock \href {https://doi.org/10.18653/v1/2021.emnlp-main.728} {Towards
  label-agnostic emotion embeddings}.
\newblock In \emph{Proceedings of the 2021 Conference on Empirical Methods in
  Natural Language Processing}, pages 9231--9249, Online and Punta Cana,
  Dominican Republic. Association for Computational Linguistics.

\bibitem[{Busso et~al.(2008)Busso, Bulut, Lee, Kazemzadeh, Mower, Kim, Chang,
  Lee, and Narayanan}]{Busso2008}
Carlos Busso, Murtaza Bulut, Chi-Chun Lee, Abe Kazemzadeh, Emily Mower, Samuel
  Kim, Jeannette~N. Chang, Sungbok Lee, and Shrikanth~S. Narayanan. 2008.
\newblock \href {https://doi.org/10.1007/s10579-008-9076-6} {Iemocap:
  interactive emotional dyadic motion capture database}.
\newblock \emph{Language Resources and Evaluation}, 42(4):335.

\bibitem[{Caiani and Di~Cocco(2023)}]{caiani_dicocco_2023}
Manuela Caiani and Jessica Di~Cocco. 2023.
\newblock \href {https://doi.org/10.1017/ipo.2023.8} {Populism and emotions: a
  comparative study using machine learning}.
\newblock \emph{Italian Political Science Review / Rivista Italiana di Scienza
  Politica}, page 1–16.

\bibitem[{Cambria et~al.(2022)Cambria, Liu, Decherchi, Xing, and
  Kwok}]{cambria-etal-2022-senticnet}
Erik Cambria, Qian Liu, Sergio Decherchi, Frank Xing, and Kenneth Kwok. 2022.
\newblock \href {https://aclanthology.org/2022.lrec-1.408} {{S}entic{N}et 7: A
  commonsense-based neurosymbolic {AI} framework for explainable sentiment
  analysis}.
\newblock In \emph{Proceedings of the Thirteenth Language Resources and
  Evaluation Conference}, pages 3829--3839, Marseille, France. European
  Language Resources Association.

\bibitem[{Campagnano et~al.(2022)Campagnano, Conia, and
  Navigli}]{Campagnano2022}
Cesare Campagnano, Simone Conia, and Roberto Navigli. 2022.
\newblock \href {https://doi.org/10.18653/v1/2022.acl-long.314} {{SRL4E} {--}
  {S}emantic {R}ole {L}abeling for {E}motions: {A} unified evaluation
  framework}.
\newblock In \emph{Proceedings of the 60th Annual Meeting of the Association
  for Computational Linguistics (Volume 1: Long Papers)}, pages 4586--4601,
  Dublin, Ireland. Association for Computational Linguistics.

\bibitem[{Casel et~al.(2021)Casel, Heindl, and Klinger}]{Casel2021}
Felix Casel, Amelie Heindl, and Roman Klinger. 2021.
\newblock \href {https://aclanthology.org/2021.konvens-1.5} {Emotion
  recognition under consideration of the emotion component process model}.
\newblock In \emph{KONVENS 2021}.

\bibitem[{Cevher et~al.(2019)Cevher, Zepf, and Klinger}]{Cevher2019}
Deniz Cevher, Sebastian Zepf, and Roman Klinger. 2019.
\newblock \href
  {https://corpora.linguistik.uni-erlangen.de/data/konvens/proceedings/papers/KONVENS2019_paper_16.pdf}
  {Towards multimodal emotion recognition in german speech events in cars using
  transfer learning}.
\newblock In \emph{KONVENS}.

\bibitem[{Chauhan et~al.(2020)Chauhan, S~R, Ekbal, and
  Bhattacharyya}]{chauhan-etal-2020-sentiment}
Dushyant~Singh Chauhan, Dhanush S~R, Asif Ekbal, and Pushpak Bhattacharyya.
  2020.
\newblock \href {https://doi.org/10.18653/v1/2020.acl-main.401} {Sentiment and
  emotion help sarcasm? a multi-task learning framework for multi-modal
  sarcasm, sentiment and emotion analysis}.
\newblock In \emph{Proceedings of the 58th Annual Meeting of the Association
  for Computational Linguistics}, pages 4351--4360, Online. Association for
  Computational Linguistics.

\bibitem[{Cortal et~al.(2023)Cortal, Finkel, Paroubek, and
  Ye}]{cortal-etal-2023-emotion}
Gustave Cortal, Alain Finkel, Patrick Paroubek, and Lina Ye. 2023.
\newblock \href {https://aclanthology.org/2023.latechclfl-1.8} {Emotion
  recognition based on psychological components in guided narratives for
  emotion regulation}.
\newblock In \emph{Proceedings of the 7th Joint SIGHUM Workshop on
  Computational Linguistics for Cultural Heritage, Social Sciences, Humanities
  and Literature}, pages 72--81, Dubrovnik, Croatia. Association for
  Computational Linguistics.

\bibitem[{Dankers et~al.(2019)Dankers, Rei, Lewis, and
  Shutova}]{dankers-etal-2019-modelling}
Verna Dankers, Marek Rei, Martha Lewis, and Ekaterina Shutova. 2019.
\newblock \href {https://doi.org/10.18653/v1/D19-1227} {Modelling the interplay
  of metaphor and emotion through multitask learning}.
\newblock In \emph{Proceedings of the 2019 Conference on Empirical Methods in
  Natural Language Processing and the 9th International Joint Conference on
  Natural Language Processing (EMNLP-IJCNLP)}, pages 2218--2229, Hong Kong,
  China. Association for Computational Linguistics.

\bibitem[{Dellagiacoma et~al.(2011)Dellagiacoma, Zontone, Boato, and
  Albertazzi}]{Dellagiacoma2011}
Michela Dellagiacoma, Pamela Zontone, Giulia Boato, and Liliana Albertazzi.
  2011.
\newblock \href {https://doi.org/10.1145/2064448.2064470} {Emotion based
  classification of natural images}.
\newblock In \emph{Proceedings of the 2011 International Workshop on DETecting
  and Exploiting Cultural DiversiTy on the Social Web}, DETECT '11, page
  17–22, New York, NY, USA. Association for Computing Machinery.

\bibitem[{Demszky et~al.(2020)Demszky, Movshovitz-Attias, Ko, Cowen, Nemade,
  and Ravi}]{Demszky2020}
Dorottya Demszky, Dana Movshovitz-Attias, Jeongwoo Ko, Alan Cowen, Gaurav
  Nemade, and Sujith Ravi. 2020.
\newblock \href {https://doi.org/10.18653/v1/2020.acl-main.372}
  {{G}o{E}motions: A dataset of fine-grained emotions}.
\newblock In \emph{Proceedings of the 58th Annual Meeting of the Association
  for Computational Linguistics}, pages 4040--4054, Online. Association for
  Computational Linguistics.

\bibitem[{Devlin et~al.(2019)Devlin, Chang, Lee, and
  Toutanova}]{devlin-etal-2019-bert}
Jacob Devlin, Ming-Wei Chang, Kenton Lee, and Kristina Toutanova. 2019.
\newblock \href {https://doi.org/10.18653/v1/N19-1423} {{BERT}: Pre-training of
  deep bidirectional transformers for language understanding}.
\newblock In \emph{Proceedings of the 2019 Conference of the North {A}merican
  Chapter of the Association for Computational Linguistics: Human Language
  Technologies, Volume 1 (Long and Short Papers)}, pages 4171--4186,
  Minneapolis, Minnesota. Association for Computational Linguistics.

\bibitem[{Dillard and Seo(2012)}]{Dillard2012}
James~Price Dillard and Kiwon Seo. 2012.
\newblock \href {https://doi.org/10.4135/9781452218410} {Affect and
  persuasion}.
\newblock In James~Price Dillard and Lijang Shen, editors, \emph{The {SAGE}
  Handbook of Persuasion: Developments in Theory and Practice}, chapter~10.
  SAGE Publications.

\bibitem[{Doan~Dang et~al.(2021)Doan~Dang, Oberl{\"a}nder, and
  Klinger}]{DoanDang2021}
Bao~Minh Doan~Dang, Laura Oberl{\"a}nder, and Roman Klinger. 2021.
\newblock \href {https://aclanthology.org/2021.konvens-1.7} {Emotion stimulus
  detection in {G}erman news headlines}.
\newblock In \emph{Proceedings of the 17th Conference on Natural Language
  Processing (KONVENS 2021)}, pages 73--85, D{\"u}sseldorf, Germany. KONVENS
  2021 Organizers.

\bibitem[{Dodds et~al.(2011)Dodds, Harris, Kloumann, Bliss, and
  Danforth}]{Dodds-2011-Hedonometrics}
Peter~Sheridan Dodds, Kameron~Decker Harris, Isabel~M. Kloumann, Catherine~A.
  Bliss, and Christopher~M. Danforth. 2011.
\newblock \href {https://doi.org/10.1371/journal.pone.0026752} {Temporal
  patterns of happiness and information in a global social network:
  Hedonometrics and twitter}.
\newblock \emph{\textsc{Plos One}}, 6(12):1--1.

\bibitem[{Doellinger et~al.(2021)Doellinger, Laukka, Högman, Bänziger,
  Makower, Fischer, and Hau}]{Doellinger2021}
Lillian Doellinger, Petri Laukka, Lennart~Björn Högman, Tanja Bänziger,
  Irena Makower, Håkan Fischer, and Stephan Hau. 2021.
\newblock \href {https://doi.org/10.3389/fpsyg.2021.708867} {Training emotion
  recognition accuracy: Results for multimodal expressions and facial micro
  expressions}.
\newblock \emph{Frontiers in Psychology}, 12.

\bibitem[{Ekman(1992)}]{Ekman1992}
Paul Ekman. 1992.
\newblock \href {https://doi.org/10.1080/02699939208411068} {An argument for
  basic emotions}.
\newblock \emph{Cognition \& emotion}, 6(3-4):169--200.

\bibitem[{Ekman and Cordaro(2011)}]{Ekman2011}
Paul Ekman and Daniel Cordaro. 2011.
\newblock \href {https://doi.org/10.1177/1754073911410740} {What is meant by
  calling emotions basic}.
\newblock \emph{Emotion Review}.

\bibitem[{Etienne et~al.(2022)Etienne, Battistelli, and
  Lecorv{\'e}}]{etienne-etal-2022-psycho}
Aline Etienne, Delphine Battistelli, and Gw{\'e}nol{\'e} Lecorv{\'e}. 2022.
\newblock \href {https://aclanthology.org/2022.lrec-1.64} {A
  (psycho-)linguistically motivated scheme for annotating and exploring
  emotions in a genre-diverse corpus}.
\newblock In \emph{Proceedings of the Thirteenth Language Resources and
  Evaluation Conference}, pages 603--612, Marseille, France. European Language
  Resources Association.

\bibitem[{Fan et~al.(2018)Fan, Shen, Jiang, Koenig, Xu, Kankanhalli, and
  Zhao}]{Fan2018}
Shaojing Fan, Zhiqi Shen, Ming Jiang, Bryan~L. Koenig, Juan Xu, Mohan~S.
  Kankanhalli, and Qi~Zhao. 2018.
\newblock \href {https://doi.org/10.1109/CVPR.2018.00785} {Emotional attention:
  A study of image sentiment and visual attention}.
\newblock In \emph{2018 IEEE/CVF Conference on Computer Vision and Pattern
  Recognition}, pages 7521--7531.

\bibitem[{Felbo et~al.(2017)Felbo, Mislove, S{\o}gaard, Rahwan, and
  Lehmann}]{felbo-etal-2017-using}
Bjarke Felbo, Alan Mislove, Anders S{\o}gaard, Iyad Rahwan, and Sune Lehmann.
  2017.
\newblock \href {https://doi.org/10.18653/v1/D17-1169} {Using millions of emoji
  occurrences to learn any-domain representations for detecting sentiment,
  emotion and sarcasm}.
\newblock In \emph{Proceedings of the 2017 Conference on Empirical Methods in
  Natural Language Processing}, pages 1615--1625, Copenhagen, Denmark.
  Association for Computational Linguistics.

\bibitem[{Feldman~Barrett(2017)}]{Barrett2017a}
Lisa Feldman~Barrett. 2017.
\newblock \href {https://doi.org/10.1093/scan/nsw154} {{The theory of
  constructed emotion: an active inference account of interoception and
  categorization}}.
\newblock \emph{Social Cognitive and Affective Neuroscience}, 12(11):1833.

\bibitem[{Gao et~al.(2017)Gao, Hu, Xu, Lin, He, Lu, and Wong}]{gao2017}
Qinghong Gao, Jiannan Hu, Ruifeng Xu, Gui Lin, Yulan He, Qin Lu, and Kam-Fai
  Wong. 2017.
\newblock \href
  {http://research.nii.ac.jp/ntcir/workshop/OnlineProceedings13/pdf/ntcir/01-NTCIR13-OV-ECA-GaoQ.pdf}
  {Overview of {NTCIR}-13 {ECA} task}.
\newblock In \emph{Proceedings of the 13th NTCIR Conference on Evaluation of
  Information Access Technologies}, pages 361--366, Tokyo, Japan.

\bibitem[{Geiser et~al.(2017)Geiser, Götz, Preckel, and
  Freund}]{geiser_states_2017}
Christian Geiser, Thomas Götz, Franzis Preckel, and Philipp~Alexander Freund.
  2017.
\newblock \href {https://doi.org/10.1027/1015-5759/a000413} {States and
  {Traits}: {Theories}, {Models}, and {Assessment}}.
\newblock \emph{European Journal of Psychological Assessment}, 33(4):219--223.

\bibitem[{Ghazi et~al.(2015)Ghazi, Inkpen, and Szpakowicz}]{ghazi2015}
Diman Ghazi, Diana Inkpen, and Stan Szpakowicz. 2015.
\newblock \href {https://doi.org/10.1007/978-3-319-18117-2_12} {Detecting
  emotion stimuli in emotion-bearing sentences}.
\newblock In \emph{International Conference on Intelligent Text Processing and
  Computational Linguistics}, pages 152--165. Springer.

\bibitem[{Gildea and Jurafsky(2000)}]{gildea-jurafsky-2000-automatic}
Daniel Gildea and Daniel Jurafsky. 2000.
\newblock \href {https://doi.org/10.3115/1075218.1075283} {Automatic labeling
  of semantic roles}.
\newblock In \emph{Proceedings of the 38th Annual Meeting of the Association
  for Computational Linguistics}, pages 512--520, Hong Kong. Association for
  Computational Linguistics.

\bibitem[{Golbeck et~al.(2011)Golbeck, Robles, and
  Turner}]{golbeck2011predicting}
Jennifer Golbeck, Cristina Robles, and Karen Turner. 2011.
\newblock \href {https://doi.org/10.1145/1979742.1979614} {Predicting
  personality with social media}.
\newblock In \emph{CHI'11 extended abstracts on human factors in computing
  systems}, pages 253--262.

\bibitem[{Goldberg(1999)}]{goldberg1999broad}
Lewis~R. Goldberg. 1999.
\newblock \href {https://ipip.ori.org/A%20broad-bandwidth%20inventory.pdf} {A
  broad-bandwidth, public domain, personality inventory measuring the
  lower-level facets of several five-factor models}.
\newblock \emph{Personality psychology in Europe}, 7(1):7--28.

\bibitem[{Hofmann et~al.(2021)Hofmann, Troiano, and
  Klinger}]{hofmann-etal-2021-emotion}
Jan Hofmann, Enrica Troiano, and Roman Klinger. 2021.
\newblock \href {https://aclanthology.org/2021.wassa-1.17} {Emotion-aware,
  emotion-agnostic, or automatic: Corpus creation strategies to obtain
  cognitive event appraisal annotations}.
\newblock In \emph{Proceedings of the Eleventh Workshop on Computational
  Approaches to Subjectivity, Sentiment and Social Media Analysis}, pages
  160--170, Online. Association for Computational Linguistics.

\bibitem[{Hofmann et~al.(2020)Hofmann, Troiano, Sassenberg, and
  Klinger}]{hofmann-etal-2020-appraisal}
Jan Hofmann, Enrica Troiano, Kai Sassenberg, and Roman Klinger. 2020.
\newblock \href {https://doi.org/10.18653/v1/2020.coling-main.11} {Appraisal
  theories for emotion classification in text}.
\newblock In \emph{Proceedings of the 28th International Conference on
  Computational Linguistics}, pages 125--138, Barcelona, Spain (Online).
  International Committee on Computational Linguistics.

\bibitem[{Hoorn and Konijn(2003)}]{hoorn_perceiving_2003}
Johan~F. Hoorn and Elly~A. Konijn. 2003.
\newblock \href {https://doi.org/10.1111/1468-5884.00225} {Perceiving and
  experiencing fictional characters: {An} integrative account: {Perceiving} and
  experiencing fictional characters}.
\newblock \emph{Japanese Psychological Research}, 45(4):250--268.

\bibitem[{Islam et~al.(2018)Islam, Kabir, Ahmed, Kamal, Wang, and
  Ulhaq}]{Islam2018-hc}
Md~Rafiqul Islam, Muhammad~Ashad Kabir, Ashir Ahmed, Abu Raihan~M Kamal, Hua
  Wang, and Anwaar Ulhaq. 2018.
\newblock \href {https://doi.org/10.1007/s13755-018-0046-0} {Depression
  detection from social network data using machine learning techniques}.
\newblock \emph{Health Inf. Sci. Syst.}, 6(1):8.

\bibitem[{Kern et~al.(2016)Kern, Park, Eichstaedt, Schwartz, Sap, Smith, and
  Ungar}]{kern_gaining_2016}
Margaret~L. Kern, Gregory Park, Johannes~C. Eichstaedt, H.~Andrew Schwartz,
  Maarten Sap, Laura~K. Smith, and Lyle~H. Ungar. 2016.
\newblock \href {https://doi.org/10.1037/met0000091} {Gaining insights from
  social media language: {Methodologies} and challenges.}
\newblock \emph{Psychological Methods}, 21(4):507--525.

\bibitem[{Khlyzova et~al.(2022)Khlyzova, Silberer, and
  Klinger}]{khlyzova-etal-2022-complementarity}
Anna Khlyzova, Carina Silberer, and Roman Klinger. 2022.
\newblock \href {https://doi.org/10.18653/v1/2022.wassa-1.1} {On the
  complementarity of images and text for the expression of emotions in social
  media}.
\newblock In \emph{Proceedings of the 12th Workshop on Computational Approaches
  to Subjectivity, Sentiment {\&} Social Media Analysis}, pages 1--15, Dublin,
  Ireland. Association for Computational Linguistics.

\bibitem[{Kim and Klinger(2018)}]{Kim2018}
Evgeny Kim and Roman Klinger. 2018.
\newblock \href {https://aclanthology.org/C18-1114} {Who feels what and why?
  annotation of a literature corpus with semantic roles of emotions}.
\newblock In \emph{Proceedings of the 27th International Conference on
  Computational Linguistics}, pages 1345--1359, Santa Fe, New Mexico, USA.
  Association for Computational Linguistics.

\bibitem[{Kim and Klinger(2019{\natexlab{a}})}]{kim-klinger-2019-analysis}
Evgeny Kim and Roman Klinger. 2019{\natexlab{a}}.
\newblock \href {https://doi.org/10.18653/v1/W19-3406} {An analysis of emotion
  communication channels in fan-fiction: Towards emotional storytelling}.
\newblock In \emph{Proceedings of the Second Workshop on Storytelling}, pages
  56--64, Florence, Italy. Association for Computational Linguistics.

\bibitem[{Kim and Klinger(2019{\natexlab{b}})}]{Kim2019}
Evgeny Kim and Roman Klinger. 2019{\natexlab{b}}.
\newblock \href {https://www.aclanthology.org/N19-1067} {Frowning {F}rodo,
  wincing {L}eia, and a seriously great friendship: Learning to classify
  emotional relationships of fictional characters}.
\newblock In \emph{Proceedings of the 2019 Conference of the North {A}merican
  Chapter of the Association for Computational Linguistics: Human Language
  Technologies, Volume 1 (Long and Short Papers)}, pages 647--653, Minneapolis,
  Minnesota. Association for Computational Linguistics.

\bibitem[{Kim et~al.(2017)Kim, Pad\'{o}, and Klinger}]{Kim2017a}
Evgeny Kim, Sebastian Pad\'{o}, and Roman Klinger. 2017.
\newblock \href {http://www.aclanthology.org/W17-2203} {Investigating the
  relationship between literary genres and emotional plot development}.
\newblock In \emph{Proceedings of the Joint SIGHUM Workshop on Computational
  Linguistics for Cultural Heritage, Social Sciences, Humanities and
  Literature}, pages 17--26, Vancouver, Canada. Association for Computational
  Linguistics.

\bibitem[{Kiritchenko et~al.(2016)Kiritchenko, Mohammad, and
  Salameh}]{Kiritchenko2016}
Svetlana Kiritchenko, Saif Mohammad, and Mohammad Salameh. 2016.
\newblock \href {https://doi.org/10.18653/v1/S16-1004} {{S}em{E}val-2016 task
  7: Determining sentiment intensity of {E}nglish and {A}rabic phrases}.
\newblock In \emph{Proceedings of the 10th International Workshop on Semantic
  Evaluation ({S}em{E}val-2016)}, pages 42--51, San Diego, California.
  Association for Computational Linguistics.

\bibitem[{Klinger and Cimiano(2013)}]{klinger-cimiano:2013:Short}
Roman Klinger and Philipp Cimiano. 2013.
\newblock \href {http://www.aclanthology.org/P13-2147} {Bi-directional
  inter-dependencies of subjective expressions and targets and their value for
  a joint model}.
\newblock In \emph{Proceedings of the 51st Annual Meeting of the Association
  for Computational Linguistics (Volume 2: Short Papers)}, pages 848--854,
  Sofia, Bulgaria. Association for Computational Linguistics.

\bibitem[{Klinger et~al.(2018)Klinger, De~Clercq, Mohammad, and
  Balahur}]{Klinger2018}
Roman Klinger, Orph{\'e}e De~Clercq, Saif Mohammad, and Alexandra Balahur.
  2018.
\newblock \href {https://doi.org/10.18653/v1/W18-6206} {{IEST}: {WASSA}-2018
  implicit emotions shared task}.
\newblock In \emph{Proceedings of the 9th Workshop on Computational Approaches
  to Subjectivity, Sentiment and Social Media Analysis}, pages 31--42,
  Brussels, Belgium. Association for Computational Linguistics.

\bibitem[{Kreuter et~al.(2022)Kreuter, Sassenberg, and
  Klinger}]{kreuter-etal-2022-items}
Anne Kreuter, Kai Sassenberg, and Roman Klinger. 2022.
\newblock \href {https://doi.org/10.18653/v1/2022.wassa-1.35} {Items from
  psychometric tests as training data for personality profiling models of
  {T}witter users}.
\newblock In \emph{Proceedings of the 12th Workshop on Computational Approaches
  to Subjectivity, Sentiment {\&} Social Media Analysis}, pages 315--323,
  Dublin, Ireland. Association for Computational Linguistics.

\bibitem[{Kuhn(2019)}]{kuhn_computational_2019}
Jonas Kuhn. 2019.
\newblock \href {https://doi.org/10.1007/s10579-019-09459-3} {Computational
  text analysis within the {Humanities}: {How} to combine working practices
  from the contributing fields?}
\newblock \emph{Language Resources and Evaluation}, 53(4):565--602.

\bibitem[{Labat et~al.(2022)Labat, Hadifar, Demeester, and
  Hoste}]{labat-etal-2022-emotional}
Sofie Labat, Amir Hadifar, Thomas Demeester, and Veronique Hoste. 2022.
\newblock \href {https://aclanthology.org/2022.wnut-1.12} {An emotional
  journey: Detecting emotion trajectories in {D}utch customer service
  dialogues}.
\newblock In \emph{Proceedings of the Eighth Workshop on Noisy User-generated
  Text (W-NUT 2022)}, pages 106--112, Gyeongju, Republic of Korea. Association
  for Computational Linguistics.

\bibitem[{Lee and Ashton(2018)}]{lee2018psychometric}
Kibeom Lee and Michael~C Ashton. 2018.
\newblock \href {https://doi.org/10.1177/1073191116659134} {Psychometric
  properties of the hexaco-100}.
\newblock \emph{Assessment}, 25(5):543--556.

\bibitem[{Li et~al.(2016)Li, Taheri, Tu, and Gimpel}]{li-etal-2016-commonsense}
Xiang Li, Aynaz Taheri, Lifu Tu, and Kevin Gimpel. 2016.
\newblock \href {https://doi.org/10.18653/v1/P16-1137} {Commonsense knowledge
  base completion}.
\newblock In \emph{Proceedings of the 54th Annual Meeting of the Association
  for Computational Linguistics (Volume 1: Long Papers)}, pages 1445--1455,
  Berlin, Germany. Association for Computational Linguistics.

\bibitem[{Li et~al.(2017)Li, Su, Shen, Li, Cao, and
  Niu}]{li-etal-2017-dailydialog}
Yanran Li, Hui Su, Xiaoyu Shen, Wenjie Li, Ziqiang Cao, and Shuzi Niu. 2017.
\newblock \href {https://aclanthology.org/I17-1099} {{D}aily{D}ialog: A
  manually labelled multi-turn dialogue dataset}.
\newblock In \emph{Proceedings of the Eighth International Joint Conference on
  Natural Language Processing (Volume 1: Long Papers)}, pages 986--995, Taipei,
  Taiwan. Asian Federation of Natural Language Processing.

\bibitem[{Liew et~al.(2016)Liew, Turtle, and Liddy}]{liew-etal-2016-emotweet}
Jasy Suet~Yan Liew, Howard~R. Turtle, and Elizabeth~D. Liddy. 2016.
\newblock \href {https://aclanthology.org/L16-1183} {{E}mo{T}weet-28: A
  fine-grained emotion corpus for sentiment analysis}.
\newblock In \emph{Proceedings of the Tenth International Conference on
  Language Resources and Evaluation ({LREC}'16)}, pages 1149--1156,
  Portoro{\v{z}}, Slovenia. European Language Resources Association (ELRA).

\bibitem[{Liu(2012)}]{liu2012sentiment}
Bing Liu. 2012.
\newblock \href {https://doi.org/10.1007/978-3-031-02145-9} {\emph{Sentiment
  analysis and opinion mining}}.
\newblock Synthesis lectures on human language technologies. Springer Nature
  Switzerland.

\bibitem[{Liu et~al.(2019)Liu, Ott, Goyal, Du, Joshi, Chen, Levy, Lewis,
  Zettlemoyer, and Stoyanov}]{Liu2019Roberta}
Yinhan Liu, Myle Ott, Naman Goyal, Jingfei Du, Mandar Joshi, Danqi Chen, Omer
  Levy, Mike Lewis, Luke Zettlemoyer, and Veselin Stoyanov. 2019.
\newblock {RoBERTa}: A robustly optimized {BERT} pretraining approach.
\newblock arXiv:1907.11692.
\newblock \url{https://arxiv.org/abs/1907.11692}.

\bibitem[{Lynn et~al.(2020)Lynn, Balasubramanian, and
  Schwartz}]{lynn-etal-2020-hierarchical}
Veronica Lynn, Niranjan Balasubramanian, and H.~Andrew Schwartz. 2020.
\newblock \href {https://doi.org/10.18653/v1/2020.acl-main.472} {Hierarchical
  modeling for user personality prediction: The role of message-level
  attention}.
\newblock In \emph{Proceedings of the 58th Annual Meeting of the Association
  for Computational Linguistics}, pages 5306--5316, Online. Association for
  Computational Linguistics.

\bibitem[{Martin and White(2005)}]{martin_language_2005}
J.~R. Martin and P.~R.~R. White. 2005.
\newblock \href {https://doi.org/10.1057/9780230511910} {\emph{The {Language}
  of {Evaluation}}}.
\newblock Palgrave Macmillan UK, London.

\bibitem[{Mohammad(2012)}]{mohammad-2012-emotional}
Saif Mohammad. 2012.
\newblock \href {https://aclanthology.org/S12-1033} {{\#}emotional tweets}.
\newblock In \emph{*{SEM} 2012: The First Joint Conference on Lexical and
  Computational Semantics {--} Volume 1: Proceedings of the main conference and
  the shared task, and Volume 2: Proceedings of the Sixth International
  Workshop on Semantic Evaluation ({S}em{E}val 2012)}, pages 246--255,
  Montr{\'e}al, Canada. Association for Computational Linguistics.

\bibitem[{Mohammad and Bravo-Marquez(2017)}]{Mohammad2017}
Saif Mohammad and Felipe Bravo-Marquez. 2017.
\newblock \href {https://doi.org/10.18653/v1/W17-5205} {{WASSA}-2017 shared
  task on emotion intensity}.
\newblock In \emph{Proceedings of the 8th Workshop on Computational Approaches
  to Subjectivity, Sentiment and Social Media Analysis}, pages 34--49,
  Copenhagen, Denmark. Association for Computational Linguistics.

\bibitem[{Mohammad et~al.(2018)Mohammad, Bravo-Marquez, Salameh, and
  Kiritchenko}]{Mohammad2018}
Saif Mohammad, Felipe Bravo-Marquez, Mohammad Salameh, and Svetlana
  Kiritchenko. 2018.
\newblock \href {https://doi.org/10.18653/v1/S18-1001} {{S}em{E}val-2018 task
  1: Affect in tweets}.
\newblock In \emph{Proceedings of The 12th International Workshop on Semantic
  Evaluation}, pages 1--17, New Orleans, Louisiana. Association for
  Computational Linguistics.

\bibitem[{Mohammad and Turney(2013)}]{Mohammad2013}
Saif Mohammad and Peter~D. Turney. 2013.
\newblock \href {https://doi.org/10.1111/j.1467-8640.2012.00460} {Crowdsourcing
  a word-emotion association lexicon.}
\newblock \emph{Computational Intelligence}, 29(3):436--465.

\bibitem[{Mohammad et~al.(2014)Mohammad, Zhu, and Martin}]{Mohammad2014}
Saif Mohammad, Xiaodan Zhu, and Joel Martin. 2014.
\newblock \href {https://doi.org/10.3115/v1/W14-2607} {Semantic role labeling
  of emotions in tweets}.
\newblock In \emph{Proceedings of the 5th Workshop on Computational Approaches
  to Subjectivity, Sentiment and Social Media Analysis}, pages 32--41,
  Baltimore, Maryland. Association for Computational Linguistics.

\bibitem[{Myers(1998)}]{Myers1998}
Isabel~Briggs Myers. 1998.
\newblock \href {https://worldcat.org/en/title/935541142} {\emph{Introduction
  to Type: A Guide to Understanding Your Results on the MBTI Instrument}}, 6th
  edition edition.
\newblock Cpp. Inc.

\bibitem[{Newman et~al.(1967)Newman, Summer, and Warren}]{NewmanSummer1967}
William~Herman Newman, Charles~Edgar Summer, and E.~Kirby Warren. 1967.
\newblock \href
  {https://books.google.de/books?redir_esc=y&id=zJcuAAAAMAAJ&focus=searchwithinvolume&q=Communication+is+an+exchange+of+facts%2C+ideas%2C+opinions%2C+or+emotions+by+two+or+more+persons.+The+exchange+is+success-+ful+only+when+mutual+understanding+re-+sults.}
  {\emph{The Process of Management: Concepts, Bahaviour, and Practice}}.
\newblock Prentice-Hall.

\bibitem[{Oberl\"ander et~al.(2020)Oberl\"ander, Reich, and
  Klinger}]{Oberlaender2020b}
Laura Oberl\"ander, Kevin Reich, and Roman Klinger. 2020.
\newblock \href {https://www.aclanthology.org/2020.peoples-1.12/}
  {Experiencers, stimuli, or targets: Which semantic roles enable machine
  learning to infer the emotions?}
\newblock In \emph{Proceedings of the Third Workshop on Computational Modeling
  of People{'}s Opinions, Personality, and Emotions in Social Media},
  Barcelona, Spain. Association for Computational Linguistics.

\bibitem[{Pang and Lee(2008)}]{pang_opinion_2008}
Bo~Pang and Lillian Lee. 2008.
\newblock \href {https://doi.org/10.1561/1500000011} {Opinion {Mining} and
  {Sentiment} {Analysis}}.
\newblock \emph{Foundations and Trends® in Information Retrieval},
  2(1–2):1--135.

\bibitem[{Pennebaker and King(1999)}]{Pennebaker1999}
James~W. Pennebaker and Laura~A. King. 1999.
\newblock \href {https://doi.org/10.1037/0022-3514.77.6.1296} {Linguistic
  styles: {Language} use as an individual difference.}
\newblock \emph{Journal of Personality and Social Psychology},
  77(6):1296--1312.

\bibitem[{Pestian et~al.(2012)Pestian, Matykiewicz, Linn-Gust, South, Uzuner,
  Wiebe, Cohen, Hurdle, and Brew}]{Pestian2012-su}
John~P. Pestian, Pawel Matykiewicz, Michelle Linn-Gust, Brett South, Ozlem
  Uzuner, Jan Wiebe, K~Bretonnel Cohen, John Hurdle, and Christopher Brew.
  2012.
\newblock \href {https://doi.org/10.4137/BII.S9042} {Sentiment analysis of
  suicide notes: A shared task}.
\newblock \emph{Biomed. Inform. Insights}, 5(Suppl 1):3--16.

\bibitem[{Pizzolli and
  Strapparava(2019)}]{pizzolli-strapparava-2019-personality}
Daniele Pizzolli and Carlo Strapparava. 2019.
\newblock \href {https://doi.org/10.18653/v1/W19-3411} {Personality traits
  recognition in literary texts}.
\newblock In \emph{Proceedings of the Second Workshop on Storytelling}, pages
  107--111, Florence, Italy. Association for Computational Linguistics.

\bibitem[{Plank and Hovy(2015)}]{plank-hovy-2015-personality}
Barbara Plank and Dirk Hovy. 2015.
\newblock \href {https://doi.org/10.18653/v1/W15-2913} {Personality traits on
  {T}witter{---}or{---}{H}ow to get 1,500 personality tests in a week}.
\newblock In \emph{Proceedings of the 6th Workshop on Computational Approaches
  to Subjectivity, Sentiment and Social Media Analysis}, pages 92--98, Lisboa,
  Portugal. Association for Computational Linguistics.

\bibitem[{Plaza-del Arco et~al.(2022)Plaza-del Arco, Mart{\'\i}n-Valdivia, and
  Klinger}]{Plazadelarco2021}
Flor~Miriam Plaza-del Arco, Mar{\'\i}a-Teresa Mart{\'\i}n-Valdivia, and Roman
  Klinger. 2022.
\newblock \href {https://aclanthology.org/2022.coling-1.592} {Natural language
  inference prompts for zero-shot emotion classification in text across
  corpora}.
\newblock In \emph{Proceedings of the 29th International Conference on
  Computational Linguistics}, pages 6805--6817, Gyeongju, Republic of Korea.
  International Committee on Computational Linguistics.

\bibitem[{Plutchik(2001)}]{Plutchik2001}
Robert Plutchik. 2001.
\newblock \href {https://www.jstor.org/stable/27857503} {The nature of emotions
  human emotions have deep evolutionary roots, a fact that may explain their
  complexity and provide tools for clinical practice}.
\newblock \emph{American Scientist}, 89(4):344--350.

\bibitem[{Pontiki et~al.(2016)Pontiki, Galanis, Papageorgiou, Androutsopoulos,
  Manandhar, AL-Smadi, Al-Ayyoub, Zhao, Qin, De~Clercq, Hoste, Apidianaki,
  Tannier, Loukachevitch, Kotelnikov, Bel, Jim{\'e}nez-Zafra, and
  Eryi{\u{g}}it}]{pontiki-etal-2016-semeval}
Maria Pontiki, Dimitris Galanis, Haris Papageorgiou, Ion Androutsopoulos,
  Suresh Manandhar, Mohammad AL-Smadi, Mahmoud Al-Ayyoub, Yanyan Zhao, Bing
  Qin, Orph{\'e}e De~Clercq, V{\'e}ronique Hoste, Marianna Apidianaki, Xavier
  Tannier, Natalia Loukachevitch, Evgeniy Kotelnikov, Nuria Bel,
  Salud~Mar{\'\i}a Jim{\'e}nez-Zafra, and G{\"u}l{\c{s}}en Eryi{\u{g}}it. 2016.
\newblock \href {https://doi.org/10.18653/v1/S16-1002} {{S}em{E}val-2016 task
  5: Aspect based sentiment analysis}.
\newblock In \emph{Proceedings of the 10th International Workshop on Semantic
  Evaluation ({S}em{E}val-2016)}, pages 19--30, San Diego, California.
  Association for Computational Linguistics.

\bibitem[{Pontiki et~al.(2015)Pontiki, Galanis, Papageorgiou, Manandhar, and
  Androutsopoulos}]{pontiki-etal-2015-semeval}
Maria Pontiki, Dimitris Galanis, Haris Papageorgiou, Suresh Manandhar, and Ion
  Androutsopoulos. 2015.
\newblock \href {https://doi.org/10.18653/v1/S15-2082} {{S}em{E}val-2015 task
  12: Aspect based sentiment analysis}.
\newblock In \emph{Proceedings of the 9th International Workshop on Semantic
  Evaluation ({S}em{E}val 2015)}, pages 486--495, Denver, Colorado. Association
  for Computational Linguistics.

\bibitem[{Pontiki et~al.(2014)Pontiki, Galanis, Pavlopoulos, Papageorgiou,
  Androutsopoulos, and Manandhar}]{pontiki-etal-2014-semeval}
Maria Pontiki, Dimitris Galanis, John Pavlopoulos, Harris Papageorgiou, Ion
  Androutsopoulos, and Suresh Manandhar. 2014.
\newblock \href {https://doi.org/10.3115/v1/S14-2004} {{S}em{E}val-2014 task 4:
  Aspect based sentiment analysis}.
\newblock In \emph{Proceedings of the 8th International Workshop on Semantic
  Evaluation ({S}em{E}val 2014)}, pages 27--35, Dublin, Ireland. Association
  for Computational Linguistics.

\bibitem[{Posner et~al.(2005)Posner, Russell, and Peterson}]{Posner2005}
Jonathan Posner, James~A. Russell, and Bradley~S. Peterson. 2005.
\newblock \href {https://doi.org/10.1017/S0954579405050340} {The circumplex
  model of affect: an integrative approach to affective neuroscience, cognitive
  development, and psychopathology.}
\newblock \emph{Development and Psychopathology}, 17(3):715--734.

\bibitem[{Preo{\c{t}}iuc-Pietro et~al.(2016)Preo{\c{t}}iuc-Pietro, Schwartz,
  Park, Eichstaedt, Kern, Ungar, and
  Shulman}]{preotiuc-pietro-etal-2016-modelling}
Daniel Preo{\c{t}}iuc-Pietro, H.~Andrew Schwartz, Gregory Park, Johannes
  Eichstaedt, Margaret Kern, Lyle Ungar, and Elisabeth Shulman. 2016.
\newblock \href {https://doi.org/10.18653/v1/W16-0404} {Modelling valence and
  arousal in {F}acebook posts}.
\newblock In \emph{Proceedings of the 7th Workshop on Computational Approaches
  to Subjectivity, Sentiment and Social Media Analysis}, pages 9--15, San
  Diego, California. Association for Computational Linguistics.

\bibitem[{Rajamanickam et~al.(2020)Rajamanickam, Mishra, Yannakoudakis, and
  Shutova}]{rajamanickam-etal-2020-joint}
Santhosh Rajamanickam, Pushkar Mishra, Helen Yannakoudakis, and Ekaterina
  Shutova. 2020.
\newblock \href {https://doi.org/10.18653/v1/2020.acl-main.394} {Joint
  modelling of emotion and abusive language detection}.
\newblock In \emph{Proceedings of the 58th Annual Meeting of the Association
  for Computational Linguistics}, pages 4270--4279, Online. Association for
  Computational Linguistics.

\bibitem[{Randall et~al.(2017)Randall, Isaacson, and Ciro}]{Randall2017}
Ken Randall, Mary Isaacson, and Carrie Ciro. 2017.
\newblock \href {https://doi.org/10.2307/26554264} {Validity and reliability of
  the myers-briggs personality type indicator: A systematic review and
  meta-analysis}.
\newblock \emph{Journal of Best Practices in Health Professions Diversity},
  10(1):1--27.

\bibitem[{Roccas et~al.(2002)Roccas, Sagiv, Schwartz, and
  Knafo}]{roccas2002big}
Sonia Roccas, Lilach Sagiv, Shalom~H Schwartz, and Ariel Knafo. 2002.
\newblock \href {https://doi.org/10.1177/0146167202289008} {The big five
  personality factors and personal values}.
\newblock \emph{Personality and social psychology bulletin}, 28(6):789--801.

\bibitem[{Roseman(2013)}]{roseman_appraisal_2013}
Ira~J. Roseman. 2013.
\newblock \href {https://doi.org/10.1177/1754073912469591} {Appraisal in the
  {Emotion} {System}: {Coherence} in {Strategies} for {Coping}}.
\newblock \emph{Emotion Review}, 5(2):141--149.

\bibitem[{Scarantino(2016)}]{Scarantino2016}
Andrea Scarantino. 2016.
\newblock \href
  {https://www.guilford.com/books/Handbook-of-Emotions/Barrett-Lewis-Haviland-Jones/9781462536368/contents}
  {The philosophy of emotions and its impact on affective science}.
\newblock In \emph{Handbook of emotions}, chapter~4, pages 3--48. Guilford
  Press New York, NY.

\bibitem[{Schaffer(1995)}]{schaffer_shocking_1995}
Deborah Schaffer. 1995.
\newblock \href {http://www.jstor.org/stable/42577612} {\textsc{Shocking
  Secrets Reveales}! {The} {Language} of {Tabloid} {Headlines}}.
\newblock \emph{ETC: A Review of General Semantics}, 52(1):27--46.
\newblock Publisher: Institute of General Semantics.

\bibitem[{Scherer et~al.(2001)Scherer, Schorr, and Johnstone}]{Scherer2001a}
Klaus~R. Scherer, Angela Schorr, and Tom Johnstone. 2001.
\newblock \href
  {https://global.oup.com/academic/product/appraisal-processes-in-emotion-9780195130072}
  {\emph{Appraisal considered as a process of multi-level sequential
  checking}}, volume~92.
\newblock Oxford University Press.

\bibitem[{Schuff et~al.(2017)Schuff, Barnes, Mohme, Pad{\'o}, and
  Klinger}]{schuff-etal-2017-annotation}
Hendrik Schuff, Jeremy Barnes, Julian Mohme, Sebastian Pad{\'o}, and Roman
  Klinger. 2017.
\newblock \href {https://doi.org/10.18653/v1/W17-5203} {Annotation, modelling
  and analysis of fine-grained emotions on a stance and sentiment detection
  corpus}.
\newblock In \emph{Proceedings of the 8th Workshop on Computational Approaches
  to Subjectivity, Sentiment and Social Media Analysis}, pages 13--23,
  Copenhagen, Denmark. Association for Computational Linguistics.

\bibitem[{Shaikh et~al.(2009)Shaikh, Prendinger, and Ishizuka}]{Shaikh2009}
Mostafa Al~Masum Shaikh, Helmut Prendinger, and Mitsuru Ishizuka. 2009.
\newblock \href {https://doi.org/10.1007/978-1-84800-306-4_4} {A linguistic
  interpretation of the {OCC} emotion model for affect sensing from text}.
\newblock \emph{Affective Information Processing}.

\bibitem[{Sinha et~al.(2015)Sinha, Dey, Mitra, and
  Basu}]{sinha-etal-2015-mining}
Priyanka Sinha, Lipika Dey, Pabitra Mitra, and Anupam Basu. 2015.
\newblock \href {https://doi.org/10.18653/v1/W15-2920} {Mining \textsc{Hexaco}
  personality traits from enterprise social media}.
\newblock In \emph{Proceedings of the 6th Workshop on Computational Approaches
  to Subjectivity, Sentiment and Social Media Analysis}, pages 140--147,
  Lisboa, Portugal. Association for Computational Linguistics.

\bibitem[{Smith and Ellsworth(1985)}]{Smith1985}
Craig.~A. Smith and Phoebe.~C. Ellsworth. 1985.
\newblock \href
  {https://www.researchgate.net/publication/19274815_Patterns_of_Cognitive_Appraisal_in_Emotion}
  {Patterns of cognitive appraisal in emotion}.
\newblock \emph{Journal of Personality and Social Psychology}, 48(4).

\bibitem[{Stajner and Yenikent(2021)}]{stajner-yenikent-2021-mbti}
Sanja Stajner and Seren Yenikent. 2021.
\newblock \href {https://doi.org/10.18653/v1/2021.eacl-main.312} {Why is {MBTI}
  personality detection from texts a difficult task?}
\newblock In \emph{Proceedings of the 16th Conference of the European Chapter
  of the Association for Computational Linguistics: Main Volume}, pages
  3580--3589, Online. Association for Computational Linguistics.

\bibitem[{Steunebrink et~al.(2009)Steunebrink, Dastani, and
  Meyer}]{Steunebrink2009}
Bas~R. Steunebrink, Mehdi Dastani, and John-Jules~Ch. Meyer. 2009.
\newblock The occ model revisited.
\newblock Online:
  \url{https://people.idsia.ch/~steunebrink/Publications/KI09_OCC_revisited.pdf}.

\bibitem[{Stranisci et~al.(2022)Stranisci, Frenda, Ceccaldi, Basile, Damiano,
  and Patti}]{Stranisci2022}
Marco~Antonio Stranisci, Simona Frenda, Eleonora Ceccaldi, Valerio Basile,
  Rossana Damiano, and Viviana Patti. 2022.
\newblock \href {https://aclanthology.org/2022.lrec-1.406} {{APPR}eddit: a
  corpus of {R}eddit posts annotated for appraisal}.
\newblock In \emph{Proceedings of the Thirteenth Language Resources and
  Evaluation Conference}, pages 3809--3818, Marseille, France. European
  Language Resources Association.

\bibitem[{Strapparava and Mihalcea(2007)}]{strapparava-mihalcea-2007-semeval}
Carlo Strapparava and Rada Mihalcea. 2007.
\newblock \href {https://aclanthology.org/S07-1013} {{S}em{E}val-2007 task 14:
  Affective text}.
\newblock In \emph{Proceedings of the Fourth International Workshop on Semantic
  Evaluations ({S}em{E}val-2007)}, pages 70--74, Prague, Czech Republic.
  Association for Computational Linguistics.

\bibitem[{Troiano et~al.(2023{\natexlab{a}})Troiano, Klinger, and
  PadÃ³}]{Troiano2023a}
Enrica Troiano, Roman Klinger, and Sebastian PadÃ³. 2023{\natexlab{a}}.
\newblock \href {https://doi.org/10.3384/nejlt.2000-1533.2023.4361} {On the
  relationship between frames and emotionality in text}.
\newblock \emph{Northern European Journal of Language Technology}, 9(1).

\bibitem[{Troiano et~al.(2022)Troiano, Oberlaender, Wegge, and
  Klinger}]{troiano-etal-2022-x}
Enrica Troiano, Laura Ana~Maria Oberlaender, Maximilian Wegge, and Roman
  Klinger. 2022.
\newblock \href {https://aclanthology.org/2022.lrec-1.146} {x-en{VENT}: A
  corpus of event descriptions with experiencer-specific emotion and appraisal
  annotations}.
\newblock In \emph{Proceedings of the Thirteenth Language Resources and
  Evaluation Conference}, pages 1365--1375, Marseille, France. European
  Language Resources Association.

\bibitem[{Troiano et~al.(2023{\natexlab{b}})Troiano, Oberl\"ander, and
  Klinger}]{Troiano2023}
Enrica Troiano, Laura Oberl\"ander, and Roman Klinger. 2023{\natexlab{b}}.
\newblock \href {https://doi.org/10.1162/coli_a_00461} {Dimensional modeling of
  emotions in text with appraisal theories: Corpus creation, annotation
  reliability, and prediction}.
\newblock \emph{Computational Linguistics}, 49(1).

\bibitem[{Troiano et~al.(2019)Troiano, Pad{\'o}, and
  Klinger}]{troiano-etal-2019-crowdsourcing}
Enrica Troiano, Sebastian Pad{\'o}, and Roman Klinger. 2019.
\newblock \href {https://doi.org/10.18653/v1/P19-1391} {Crowdsourcing and
  validating event-focused emotion corpora for {G}erman and {E}nglish}.
\newblock In \emph{Proceedings of the 57th Annual Meeting of the Association
  for Computational Linguistics}, pages 4005--4011, Florence, Italy.
  Association for Computational Linguistics.

\bibitem[{Trouillon et~al.(2017)Trouillon, Dance, Gaussier, Welbl, Riedel, and
  Bouchard}]{Trouillon2017}
Th\'{e}o Trouillon, Christopher~R. Dance, \'{E}ric Gaussier, Johannes Welbl,
  Sebastian Riedel, and Guillaume Bouchard. 2017.
\newblock \href {https://doi.org/10.5555/3122009.3208011} {Knowledge graph
  completion via complex tensor factorization}.
\newblock \emph{Journal Machine Learning Research}, 18(1):4735–4772.

\bibitem[{Udochukwu and He(2015)}]{Udochukwu2015}
Orizu Udochukwu and Yulan He. 2015.
\newblock \href
  {https://www.springerprofessional.de/en/a-rule-based-approach-to-implicit-emotion-detection-in-text/2464810}
  {A rule-based approach to implicit emotion detection in text}.
\newblock In \emph{Natural Language Processing and Information Systems}.

\bibitem[{Verhoeven et~al.(2016)Verhoeven, Daelemans, and
  Plank}]{verhoeven-etal-2016-twisty}
Ben Verhoeven, Walter Daelemans, and Barbara Plank. 2016.
\newblock \href {https://aclanthology.org/L16-1258} {{T}wi{S}ty: A multilingual
  {T}witter stylometry corpus for gender and personality profiling}.
\newblock In \emph{Proceedings of the Tenth International Conference on
  Language Resources and Evaluation ({LREC}'16)}, pages 1632--1637,
  Portoro{\v{z}}, Slovenia. European Language Resources Association (ELRA).

\bibitem[{\v{S}tajner and Klinger(2023)}]{StajnerKlinger2023}
Sanja \v{S}tajner and Roman Klinger. 2023.
\newblock \href {https://aclanthology.org/2023.eacl-tutorials.2} {Emotion
  analysis from texts}.
\newblock In \emph{Proceedings of the 17th Conference of the European Chapter
  of the Association for Computational Linguistics: Tutorial Abstracts}, pages
  7--12, Dubrovnik, Croatia. Association for Computational Linguistics.

\bibitem[{Wang et~al.(2012)Wang, Chen, Thirunarayan, and
  Sheth}]{wang_harnessing_2012}
Wenbo Wang, Lu~Chen, Krishnaprasad Thirunarayan, and Amit~P. Sheth. 2012.
\newblock \href {https://doi.org/10.1109/SocialCom-PASSAT.2012.119} {Harnessing
  {Twitter} `{Big} {Data}' for {Automatic} {Emotion} {Identification}}.
\newblock In \emph{2012 ASE/IEEE International Conference on Social Computing
  and 2012 ASE/IEEE International Conference on Privacy, Security, Risk and
  Trust}, pages 587--592. IEEE.

\bibitem[{Wegge and Klinger(2023)}]{Wegge2023}
Maximilian Wegge and Roman Klinger. 2023.
\newblock \href {https://www.romanklinger.de/publications/WeggeKlinger2023.pdf}
  {Automatic emotion experiencer recognition}.
\newblock In \emph{3rd Workshop on Computational Linguistics for the Political
  and Social Sciences (CPSS)}.

\bibitem[{Wegge et~al.(2022)Wegge, Troiano, Oberlaender, and
  Klinger}]{Wegge2022}
Maximilian Wegge, Enrica Troiano, Laura Ana~Maria Oberlaender, and Roman
  Klinger. 2022.
\newblock \href {https://aclanthology.org/2022.nlpcss-1.3}
  {Experiencer-specific emotion and appraisal prediction}.
\newblock In \emph{Proceedings of the Fifth Workshop on Natural Language
  Processing and Computational Social Science (NLP+CSS)}, pages 25--32, Abu
  Dhabi, UAE. Association for Computational Linguistics.

\bibitem[{Wiebe et~al.(2004)Wiebe, Wilson, Bruce, Bell, and
  Martin}]{wiebe-etal-2004-learning}
Janyce Wiebe, Theresa Wilson, Rebecca Bruce, Matthew Bell, and Melanie Martin.
  2004.
\newblock \href {https://doi.org/10.1162/0891201041850885} {Learning subjective
  language}.
\newblock \emph{Computational Linguistics}, 30(3):277--308.

\bibitem[{Xia and Ding(2019)}]{xia-ding-2019-emotion}
Rui Xia and Zixiang Ding. 2019.
\newblock \href {https://doi.org/10.18653/v1/P19-1096} {Emotion-cause pair
  extraction: A new task to emotion analysis in texts}.
\newblock In \emph{Proceedings of the 57th Annual Meeting of the Association
  for Computational Linguistics}, pages 1003--1012, Florence, Italy.
  Association for Computational Linguistics.

\bibitem[{Zhou et~al.(2016)Zhou, Shi, Tian, Qi, Li, Hao, and
  Xu}]{zhou-etal-2016-attention}
Peng Zhou, Wei Shi, Jun Tian, Zhenyu Qi, Bingchen Li, Hongwei Hao, and Bo~Xu.
  2016.
\newblock \href {https://doi.org/10.18653/v1/P16-2034} {Attention-based
  bidirectional long short-term memory networks for relation classification}.
\newblock In \emph{Proceedings of the 54th Annual Meeting of the Association
  for Computational Linguistics (Volume 2: Short Papers)}, pages 207--212,
  Berlin, Germany. Association for Computational Linguistics.

\end{thebibliography}
\bibliographystyle{acl_natbib}

\appendix

\clearpage
\onecolumn

\section{Visualization of the Relations Between Tasks}
\vspace{5mm}

\begingroup %
\centering
\rotatebox{270}{\includegraphics{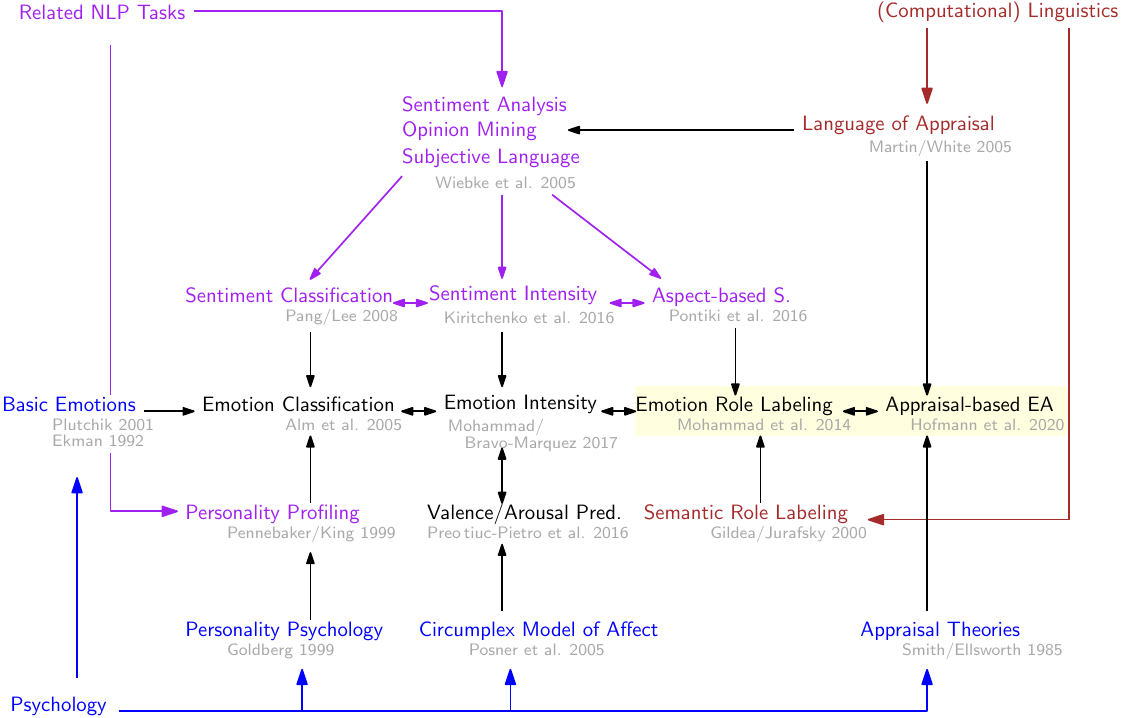}}\\[\baselineskip]%
\captionof{figure}{Visualization of relations between emotion analysis
  and other previously established tasks and studies. The
  bibliographic references are examples for the respective tasks and
  are not supposed to suggest completeness. Please see the text for a
  more comprehensive picture.}
\label{fig:sketch}
\endgroup

\end{document}